\documentclass[10pt,twocolumn,letterpaper]{article}

\usepackage{cvpr}              % To produce the CAMERA-READY version
\usepackage{tabularx}
\usepackage{amsmath}
\usepackage{makecell}
\usepackage{array}

\DeclareMathOperator*{\argtopk}{arg\,top-K}

\newcommand{\method}{\texttt{\textbf{BYOV}}\xspace}
\newcommand{\bVego}{\mathbf{V}_\text{ego}}
\newcommand{\bvego}{\mathbf{v}_\text{ego}}
\newcommand{\bVexo}{\mathbf{V}_\text{exo}}
\newcommand{\bvexo}{\mathbf{v}_\text{exo}}
\newcommand{\bZego}{\mathbf{Z}_\text{ego}}
\newcommand{\bZexo}{\mathbf{Z}_\text{exo}}

\newcommand{\Tego}{T_\text{ego}}
\newcommand{\Texo}{T_\text{exo}}
\newcommand{\Tegomsm}{T^{\text{MSM}}_\text{ego}}
\newcommand{\Texomsm}{T^{\text{MSM}}_\text{exo}}
\newcommand{\Tegomcm}{T^{\text{MCM}}_\text{ego}}
\newcommand{\Texomcm}{T^{\text{MCM}}_\text{exo}}

\newcommand{\bXego}{\mathbf{X}_\text{ego}}
\newcommand{\bXexo}{\mathbf{X}_\text{exo}}
\newcommand{\bxego}{\mathbf{x}_\text{ego}}
\newcommand{\bxexo}{\mathbf{x}_\text{exo}}
\newcommand{\bx}{\mathbf{x}}

\newcommand{\bbXego}{\bar{\mathbf{X}}_\text{ego}}
\newcommand{\bbXexo}{\bar{\mathbf{X}}_\text{exo}}

\newcommand{\bYego}{\mathbf{Y}_\text{ego}}
\newcommand{\bYexo}{\mathbf{Y}_\text{exo}}

\newcommand{\paragrapht}[1]{\noindent\textbf{#1}}  % tidy \paragraph

\usepackage{pifont}
\usepackage{amssymb}

\newcommand{\cmark}{\ding{51}}%
\newcommand{\xmark}{\ding{55}}%

\newcommand\blfootnote[1]{%
  \begingroup
  \renewcommand\thefootnote{}\footnote{#1}%
  \addtocounter{footnote}{-1}%
  \endgroup
}

\definecolor{cvprblue}{rgb}{0.21,0.49,0.74}
\usepackage[pagebackref,breaklinks,colorlinks,allcolors=cvprblue]{hyperref}
\usepackage{multirow}
\usepackage{makecell}
\usepackage{lipsum}
\usepackage{booktabs}
\usepackage{color}
\usepackage{colortbl}
\usepackage{subcaption}
\usepackage[capitalize]{cleveref}
\crefname{section}{Sec.}{Secs.}
\Crefname{section}{Section}{Sections}
\Crefname{table}{Table}{Tables}
\crefname{table}{Tab.}{Tabs.}
\def\paperID{17417} % *** Enter the Paper ID here
\def\confName{CVPR}
\def\confYear{2025}

\title{Bootstrap Your Own Views: Masked Ego-Exo Modeling \\ for Fine-grained View-invariant Video Representations}

\author{Jungin Park$^{1}$
\quad\quad\quad Jiyoung Lee$^{2,3*}$ 
\quad\quad\quad Kwanghoon Sohn$^{1,4*}$   \vspace{5pt}\\
$^1$Yonsei University \quad\quad $^2$Ewha Womans University \\
\quad\quad $^3$NAVER AI Lab \quad\quad $^4$Korea Institute of Science and Technology (KIST)\vspace{3pt}\\
{\tt\small $\lbrace$newrun, khsohn$\rbrace$@yonsei.ac.kr} \quad\quad\quad
\tt\small lee.jiyoung@ewha.ac.kr}

\begin{document}
\maketitle
\begin{abstract}
    \blfootnote{\hskip -0.2in $*$ Corresponding authors.}
    \blfootnote{\hskip -0.2in This work was partly supported by the National Research Foundation of Korea (NRF) grant funded by the Korea government (MSIT) (No.~RS-2025-00515741) and the Yonsei Signature Research Cluster Program of 2024 (2024-22-0161).}
    View-invariant representation learning from egocentric (first-person, ego) and exocentric (third-person, exo) videos is a promising approach toward generalizing video understanding systems across multiple viewpoints.
    However, this area has been underexplored due to the substantial differences in perspective, motion patterns, and context between ego and exo views.    
    In this paper, we propose a novel masked ego-exo modeling that promotes both causal temporal dynamics and cross-view alignment, called Bootstrap Your Own Views (\method), for fine-grained view-invariant video representation learning from unpaired ego-exo videos.
    We highlight the importance of capturing the compositional nature of human actions as a basis for robust cross-view understanding.
    Specifically, self-view masking and cross-view masking predictions are designed to learn view-invariant and powerful representations concurrently.
    Experimental results demonstrate that our \method significantly surpasses existing approaches with notable gains across all metrics in four downstream ego-exo video tasks.
    The code is available at \url{https://github.com/park-jungin/byov}.

    % %% From GPT
    % Understanding actions from both egocentric (first-person) and exocentric (third-person) views is essential for building video representations applicable to various real-world tasks, including robotics and augmented reality. Prior approaches often rely on hand-object interaction (HOI) networks, which can be computationally demanding and may struggle in complex scenes without clear hand-object interactions. We propose a method to achieve fine-grained view-invariant video representations from unpaired egocentric and exocentric videos without needing an explicit HOI network. Our approach selectively focuses on motion-based features, capturing action-relevant information, and utilizes a novel masking strategy to enhance temporal consistency and view alignment. Experimental results show that our method outperforms state-of-the-art models on fine-grained tasks across regular and cross-view settings, demonstrating its effectiveness and applicability in diverse video understanding scenarios.
    % %%
    
    % Aligning egocentric and exocentric viewpoints of human activities is a key component for many applications in robotics and augmented reality.
\end{abstract}    
\section{Introduction}
\label{sec:intro}

    When babies observe the actions of others $-$such as parents, siblings, or caregivers$-$, they try to replicate those actions from their own perspective.
    It is a fundamental capability of the human cognitive system~\cite{learnfromdemon, watchandlearn}, known as \textit{observational learning}.
    This learning arises from the recognition of natural action changes from an exocentric (\ie, third-person, shortly exo) perspective to an egocentric (\ie, first-person, shortly ego) perspective.
    The skill to recognize the same action across different viewpoints is a crucial requirement in practice applications including robotics and augmented reality.
    For instance, it allows robots to understand human actions from various angles, and enables better human-robot interaction.

    \begin{figure}[t]
      \centering
        {\includegraphics[width=0.98\linewidth]{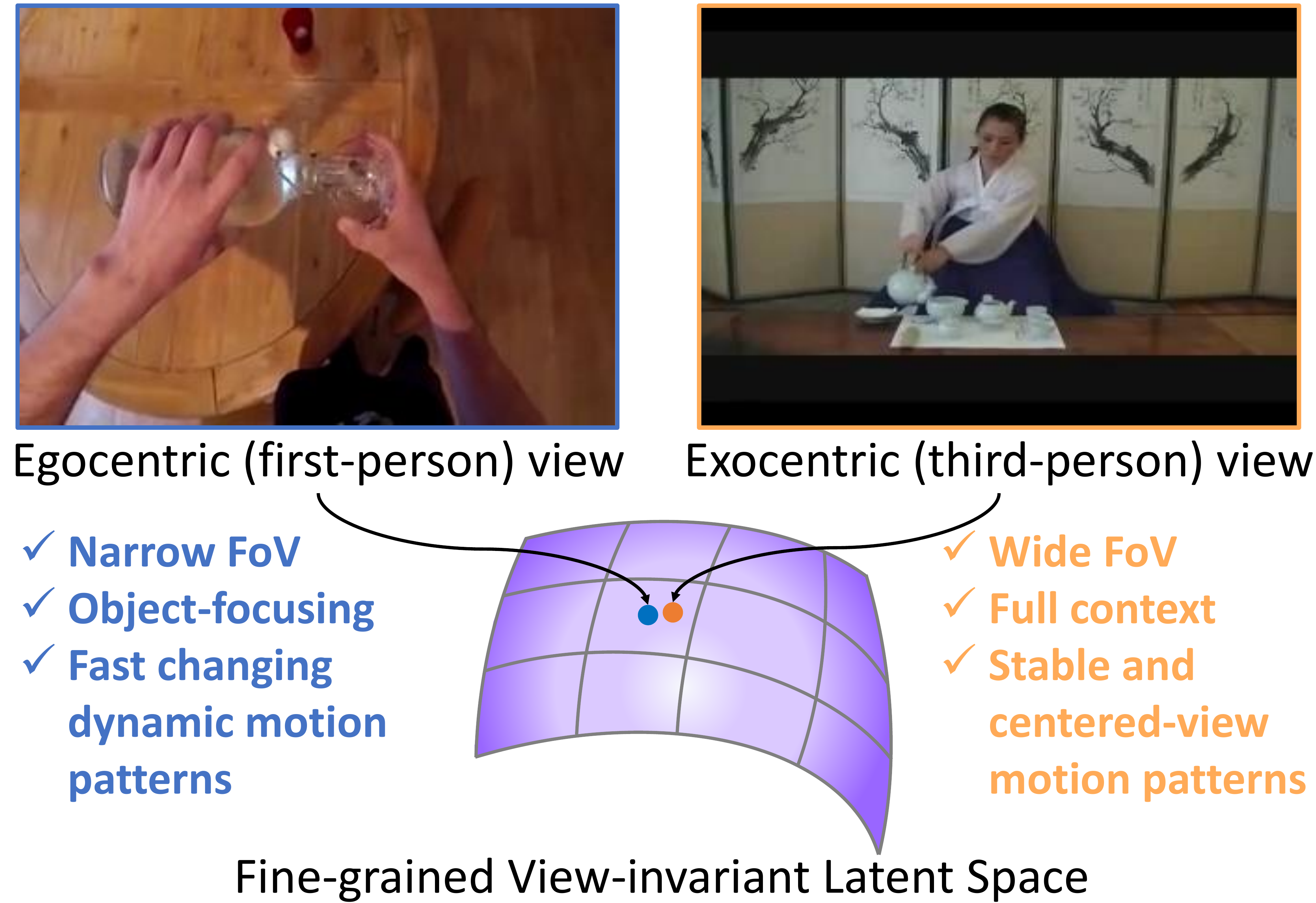}}\vspace{-3pt}       
      \caption{Challenges for view-invariant video representation learning from unpaired ego and exo videos. There is a fundamental gap between ego-exo views, such as perspective differences following the camera angle, context cues, scale and depth variations, and different motion patterns even when doing the activity.}\vspace{-3pt}
      \label{fig:1}
    \end{figure}

    However, learning view-invariant representations from both ego and exo videos poses significant challenges for the following reasons;
    1) The action videos taken simultaneously with ego-exo views are hard to control and collect, often impractical in real-world settings. 
    Therefore, we might use unpaired ego-exo videos for flexible data collection.
    2) There are fundamental differences in ego-exo videos in field-of-view (FoV) and scale variations, different focal points and motion patterns, and context about the environment, as illustrated in \cref{fig:1}.
    In ego views, objects or actions occur closer to the camera, views are dynamically changing, and the subject often focuses on what they are directly interacting with.
    However, exo views are typically more stable in moving, capturing the scene from an outside angle to contain the overall background.
    To address those challenges, prior work~\cite{ego-exo} leveraged unpaired ego-exo video data for training, where videos are annotated with the same action class but lack temporal alignment.
    However, it is limited to learning clip-level coarse representations, which constrains the ability to understand sophisticated fine-grained actions in cross-view videos.
    Recent methods~\cite{ae2,suml} have demonstrated promising progress by utilizing additional hand-object interaction detectors~\cite{ae2} or manually crafted text annotations for specific actions~\cite{suml}.
    It restricts the scalability, especially for diverse action classes.
    
    In this paper, we introduce a novel method, termed as \textit{Bootstrap Your Own Views} (\method), for fine-grained view-invariant video representation learning.
    We argue that comprehension of the accommodation of the inherent compositionally of human actions is an essential point for learning fine-grained and view-invariant representations from unpaired ego-exo videos.
    To achieve this goal, we propose masked ego-exo modeling in which the model learns temporal dynamics and aligns representations across view discrepancies.
    First, a self-view masking prediction strategy is employed by predicting masked frame embeddings based on past embeddings in each view, capturing the temporal dependencies of fine-grained actions and events.
    In addition, we propose a cross-view masking strategy, where a large proportion of embeddings is masked in one view, while the visible embeddings from the other view are used to predict the masked ones.
    This encourages the learning of consistent representations despite variations across views.

    % \begin{figure}[t]
    %   \centering
    %     \begin{subfigure}{0.9\linewidth}
    %     \centering
    %     {\includegraphics[width=0.98\linewidth]{fig/fig1-a-rev.pdf}}
    %     \caption{Temporal dependency}\label{fig:2a}
    %    \end{subfigure}  \\
    %    \begin{subfigure}{0.9\linewidth}
    %     \centering
    %     {\includegraphics[width=0.98\linewidth]{fig/fig1-b-rev.pdf}}
    %     \caption{View-invariance}\label{fig:2b}
    %    \end{subfigure}\vspace{-3pt}
    %   \caption{Motivation of \method. The pivotal role of fine-grained view-invariant video representations is to capture (a) `temporal dependency' between video frames and (b) `view-invariance' across views. We achieve this goal by the proposed self-view and cross-view masked modeling.}\vspace{-7pt}
    %   \label{fig:fig2}
    % \end{figure}
    
    We evaluate the robustness of \method in ego-exo benchmark~\cite{ae2} for four downstream fine-grained action tasks: action phase classification, frame retrieval, phase progression, and Kendall's $\tau$ for temporal alignment.
    Surprisingly, our \method demonstrates significantly superior performance, outperforming existing methods by a large margin.
    In particular, our approach shows a remarkable improvement compared to the previous SoTA~\cite{ae2}. 
    While we achieve a performance boost of over 10\% in classification, our method outperforms with a margin of 3.78 mAP@10 for frame retrieval, and a margin of 0.3424 in phase progression. 
    Additionally, in terms of Kendall's Tau, our method demonstrates a significant gap of 0.3135, further proving its superiority across various evaluation metrics. 
    These results highlight the effectiveness and robustness of the representations from our \method in diverse tasks.

    The contributions are summarized as follows:
    \begin{itemize}
        \item We introduce a novel approach, Bootstrap Your Own Views (\method), to learn fine-grained and consistent representations from unpaired and asynchronous ego-exo videos.
        \item \method builds on a masked ego-exo modeling that enables the model to learn both causal temporal dynamics and cross-view alignment.
        \item \method achieves SoTA performance by a significant margin across several downstream tasks, demonstrating the superiority of our learned representations.
    \end{itemize}

\section{Related Work}
\label{sec:2}

%-------------------------------------------------------------------------
\paragrapht{Fine-grained action recognition}
    has gained significant attention in distinguishing between subtle variations in actions that often involve similar movements or interactions. 
    Previous works for exocentric video understanding have recognized fine-grained actions by compositions of components, using the combinations of sub-actions~\cite{piergiovanni2020differentiable,li2022weakly}, words with language knowledge~\cite{doughty2022you,mettes2021object,ji2020action}, or specified attributes~\cite{rohrbach2016recognizing,zellers2017zero,zhang2021temporal}.
    Otherwise, EPIC-Kitchens~\cite{damen2018scaling} and Ego4D~\cite{ego4d} have been released as valuable benchmarks in egocentric activity understanding.
    Ego models for fine-grained action recognition have relied on temporal aggregation with global frame featrues~\cite{epicfusion,egoclip,egovlpv2} or hand-object interactions~\cite{kapidis2019egocentric, liu2022joint, tekin2019h+}.
    % While some works~\cite{epicfusion,egoclip,egovlpv2} have focused on temporal aggregation with global frame features, some works~\cite{kapidis2019egocentric, liu2022joint, tekin2019h+} have relied on analysis of hand and object interactions.
    However, those approaches trained on ego videos are limited in handling extreme variations in viewpoints.
    To tackle this problem, 
    view-invariant learning~\cite{yu2019see,ardeshir2018exocentric} has explored recognizing action in both ego and exo videos, but they require simultaneously captured ego-exo videos~\cite{egoexo4d}.
    Meanwhile, Ego-Exo~\cite{ego-exo} and AE2~\cite{ae2}, which are highly related to our work, have been proposed for reliable representation learning from unpaired ego-exo videos.
    Although Ego-Exo~\cite{ego-exo} has introduced a promising exo-to-ego transfer learning framework, it relies on the pretrained models~\cite{slowfast} to augment pseudo labels.
    The most recent work is AE2~\cite{ae2}, introducing a new benchmark for fine-grained cross-view activity understanding.
    However, an ego-exo temporal alignment framework proposed in AE2 heavily relies on off-the-shelf hand-object interaction detectors~\cite{shan2020understanding} with a classical alignment objective, \ie dynamic time warping (DTW)~\cite{dtw}, which limits to computing the optimal alignment between unpaired different view videos.
    \begin{figure*}[t]
      \centering
        \begin{subfigure}{0.23\linewidth}
        \centering
        {\includegraphics[width=0.98\linewidth]{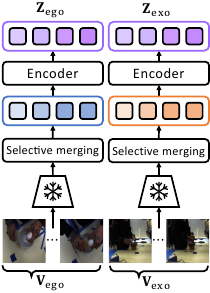}}
        \caption{Representation of \method}\label{fig:3a}
       \end{subfigure}  \hfill
       \begin{subfigure}{0.23\linewidth}
        \centering
        {\includegraphics[width=0.98\linewidth]{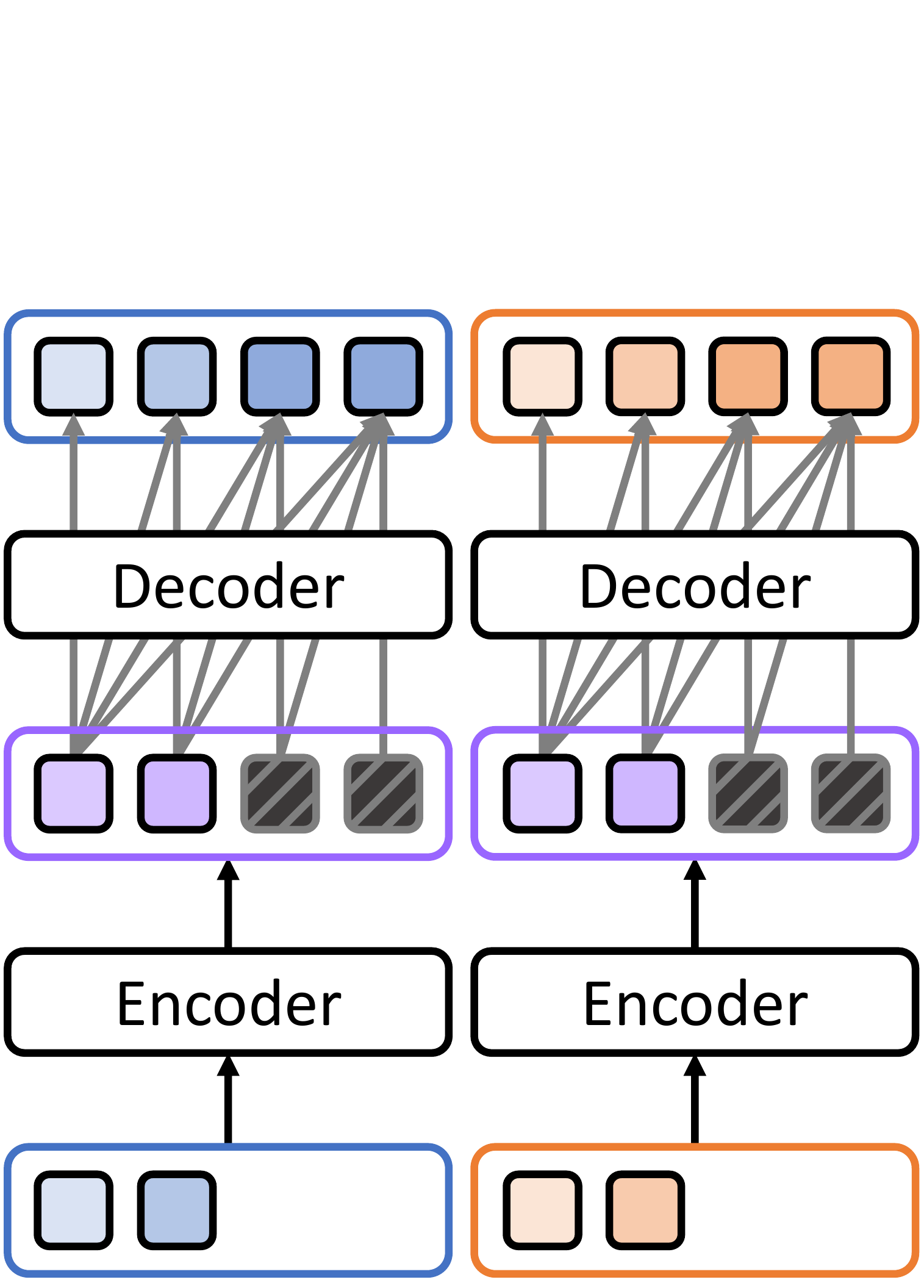}}
        \caption{Masked self-view modeling}\label{fig:3b}
       \end{subfigure}  \hfill
       \begin{subfigure}{0.48\linewidth}
        \centering
        {\includegraphics[width=0.98\linewidth]{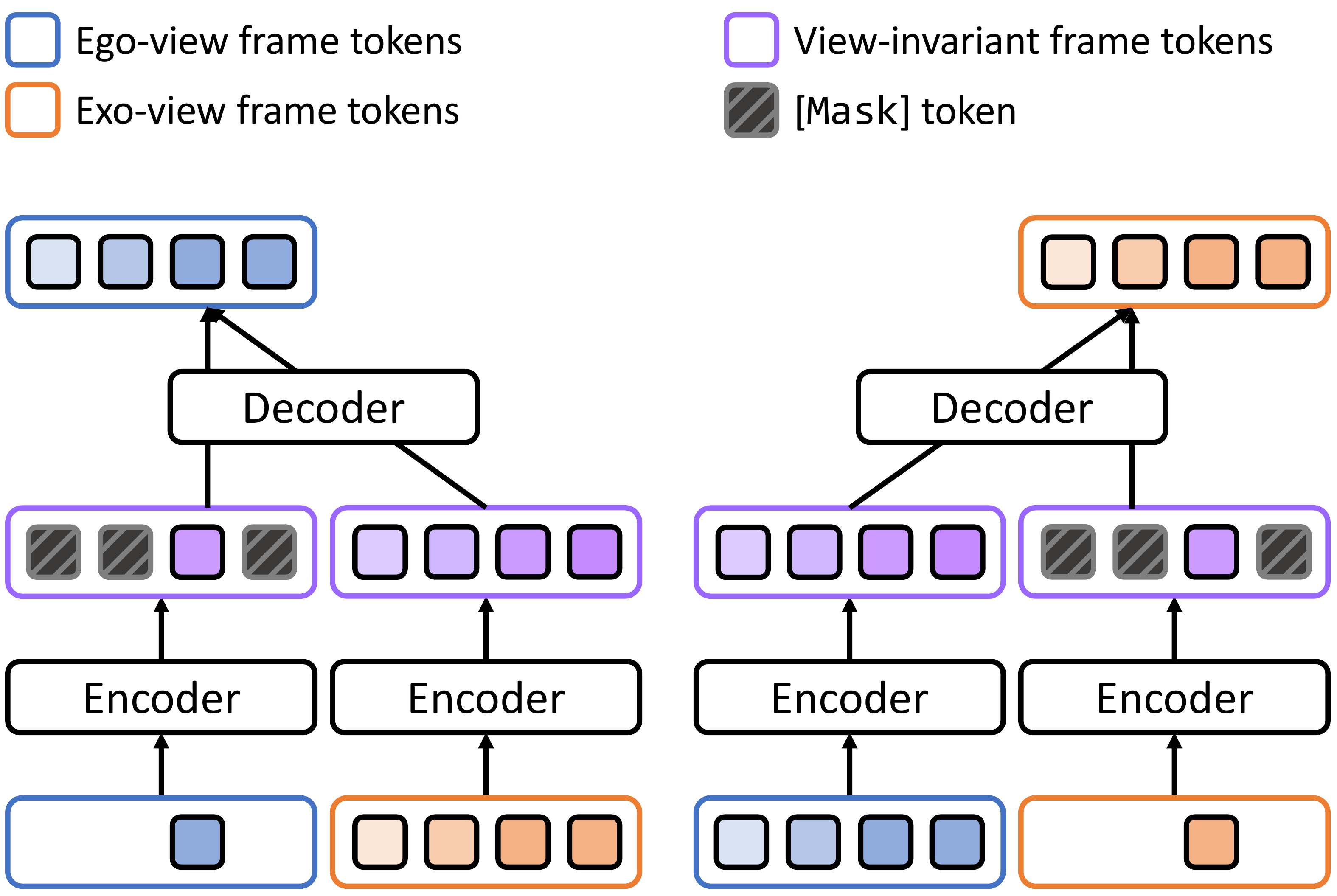}}
        \caption{Masked cross-view modeling}\label{fig:3c}
       \end{subfigure}
       \vspace{-3pt}
      \caption{The overall framework of \method. The proposed masked (b) self-view and (c) cross-view modeling encourages learning fine-grained view-invariant representations from unpaired ego and exo videos. We employ an encoder-decoder framework across disparate views, which is trained simultaneously to predict both frame tokens from its own view and frame tokens from a different view performing the same action. As shown in (a), we note that the decoder is discarded in performing downstream tasks.}%\vspace{-3pt}
      \label{fig:short}
    \end{figure*}

\paragrapht{Masked modeling in videos}, inspired by the success of masked language models such as BERT~\cite{bert}, has recently shown promise for learning rich spatiotemporal representations from videos in a self-supervised manner. 
    Commonly, a transformer-based encoder-decoder architecture~\cite{mae} is used to reconstruct masked portions of frame sequences.
    The encoder captures contextual information to infer missing content in the decoder.
    For instance, VideoMAE~\cite{videomae} employs masked autoencoding by making video patches across both temporal and spatial views.
    Instead, MaskFeat~\cite{maskfeat} performs the masked modeling to directly regress the features (\eg, HOG~\cite{hog}).
    Meanwhile, some works~\cite{beit, maskvit} have used a discrete variational autoencoder (dVAE) to compress frames into smaller visual tokens.
    Contrary to the works for exocentric video~\cite{masked_spatiotemporal, maskfeat, vimpac, videomae}, masked modeling has also been explored to learn egocentric video representations in a data-efficient manner~\cite{ego-only}.
    While the method has shown strong transferability across egocentric video tasks, such as action recognition and robot manipulation, it has only been applied to single-view ego videos. 
    As a result, the effectiveness of masked modeling for cross-view video representation learning remains largely unexplored.

\section{Bootstrap Your Own Views}
\label{sec:3}

    \subsection{Overview of \method}
    Given unpaired and asynchronous ego and exo videos (\ie, recorded independently, but belonging to the same action class), our primary goal is to learn the temporal dependency of sophisticated action and view invariance features.
    To this end, we incorporate robust masked visual modeling~\cite{mae} into fine-grained view-invariant video representation learning.
    In contrast to previous works~\cite{mae, videomae} that predict masked patches in self-supervision only, our \method employs two different masked modeling methods according to the learning objects.
    First, masked self-view modeling reconstructs frame-level token embeddings from the own view video to capture the temporal dependency of fine-grained activity between dense frames.
    Secondly, masked cross-view modeling learns view-invariant temporal features by predicting the different view's latents from one another.
    
    \method consists of three modules: 
    1) a pre-trained image encoder to extract frame token embeddings; 
    2) a transformer encoder to map frame embeddings to a view-invariant latent space; and 
    3) a transformer decoder to reconstruct masked frame embeddings.
    Specifically, vision transformers~\cite{vit, swin} compute patch-wise features for each frame, but it increases the computation demands to process all frames in the video.
    Given that frames in the video containing fine-grained activity are slightly different in spatial regions, we introduce a frame token embedding selection approach to efficiently estimate global frame token embeddings.
    For the transformer-based encoder and decoder, we adopt an asymmetric design following the previous works~\cite{mae, videomae} where the encoder processes only the partial frame token embeddings, while the decoder reconstructs the whole frame token embeddings from the view-invariant embeddings and mask token embeddings.
    After pretraining, the decoder is discarded and the encoder is used to project each frame from ego and exo videos into the shared view-invariant latent space.
    In the following sections, we present \method in detail.
    % first formulate the problem~\secref{sec:formulation} and present \method 
    
% \subsection{Problem formulation}\label{sec:formulation}
%     Let $f_{\theta}(\cdot)$ represent a pre-trained image encoder (i.e., ViT~\cite{vit}) parameterized by $\theta$, $g_{\phi}(\cdot)$, and $h_{\psi}(\cdot)$ are an encoder and a decoder of an autoencoder parameterized by $\phi$ and $\psi$, respectively.
%     Given an ego video $\bVego = \{\bvego^t\}_{t=1}^{\Tego}$ with $\Tego$ frames and an exo video $\bVexo = \{\bvexo^t\}_{t=1}^{\Texo}$ with $\Texo$ frames, we first encode each frame 
    
% \subsection{Masked ego-exo modeling}\label{sec:byov}
    \subsection{Frame encoding}
    Let $f_{\theta}(\cdot)$ represent an image encoder (\eg, ViT~\cite{vit}) pretrained from image datasets, parameterized by $\theta$, $\bVego = \{\bvego^t\}_{t=1}^{\Tego}$ and $\bVexo = \{\bvexo^t\}_{t=1}^{\Texo}$ are an ego and exo video composing sequence of $\Tego$ and $\Texo$ frames, respectively.
    For each view video, we first encode frames with $f_\theta(\cdot)$ into $d$-dimensional embedding space to obtain $N$ token embeddings for each frame, such that,
    \begin{equation}
    \begin{aligned}
        \bXego = \{\bxego^t\}_{t=1}^{\Tego} = f_\theta (\bVego) \in \mathbb{R}^{\Tego \times N \times d}, \\
        \bXexo = \{\bxexo^t\}_{t=1}^{\Texo} = f_\theta (\bVexo) \in \mathbb{R}^{\Texo \times N \times d}.
        % \bXego = \{\bxego^t | \bxego^t \in \mathbb{R}^{N\times d}, t=1,...,\Tego \}, \\
        % \bXexo = \{\bxexo^t | \bxexo^t \in \mathbb{R}^{N\times d}, t=1,...,\Texo \}.
    \end{aligned}
    \end{equation}
    To reduce the training cost, we keep the image encoder frozen during training.
    Even though image embeddings from the pretrained image encoder have robust representative power, they are often limited to capturing temporal dynamics~\cite{dualpath}.
    We thereby develop \method to use those embeddings in the ego-exo representation learning with videos.
    
    % We normalize each token embedding with $\ell$-2 normalization:
    % \begin{equation}
    %     \begin{aligned}
    %         \bxego^t \leftarrow \bxego^t / ||\bxego^t||_2, \quad \bxexo^t \leftarrow \bxexo^t / ||\bxexo^t||_2
    %     \end{aligned}
    % \end{equation}
    % Note that, we keep the image encoder frozen during training.

    \subsection{Selective token merging}
    Typically views representing atomic human actions are visually similar, and show only slight local differences (typically around a hand-object interaction region) across frames.
    For this reason, fine-grained representations have to be captured in ego-exo videos.
    While AE2~\cite{ae2} has employed hand-object interaction detector~\cite{shan2020understanding} to capture action-related regions, some ego works~\cite{egoexolearn, plizzari2022e2} have incorporated additional temporal dynamic signals such as optical flows and eye gaze signals.
    However, they have increased the burden of computational costs.
     
    Instead, we propose a simple but effective alternative solution that selects a set of tokens based on the difference between token embeddings of consecutive frames.
    Since local patches of each frame are encoded into each token embedding in the vision transformer, the difference between token embeddings located in the same position but from different frames can naively reflect the transition of local regions over time~\cite{choi2024vid}.
    Specifically, we compute the absolute difference value of the token embeddings between two consecutive frames and select top-$K$ tokens with large values:
    % The sets of selected tokens corresponding to each view video $\bXego'$ and $\bXexo'$ can be represented as: (REWRITE)
    \begin{equation}
    \begin{aligned}
        n^t &= \argtopk (\mathbf{s}^t), \\
        \text{where} \quad \mathbf{s}^t &= \frac{1}{d} \sum_d |\bx^t - \bx^{t+1}| \in \mathbb{R}^{N},
    \end{aligned}
    \end{equation}
    where $n^t$ indicates the indices of the selected tokens in each frame.
    The selected $K$ token embeddings are averaged to obtain a set of frame token embeddings $\bbXego \in \mathbb{R}^{\Tego \times d}$ and $\bbXexo \in \mathbb{R}^{\Texo \times d}$, which are used as the masked-target and encoded token embeddings to be reconstructed by the encoder $g_\phi(\cdot)$ and the decoder $h_\psi(\cdot)$.
    % \begin{equation}
    % \begin{aligned}
    %     \bbXego = \{\bbxego^t | \bbxego^t = \frac{1}{K}\bxego'^t, t=1,...,\Tego\} \in \mathbb{R}^{\Tego \times d},  \\
    %     \bbXexo = \{\bbxexo^t | \bbxexo^t = \frac{1}{K}\bxexo'^t, t=1,...,\Texo\} \in \mathbb{R}^{\Texo \times d}.        
    % \end{aligned}
    % \end{equation}
    % The frame embeddings are used as the target embeddings to be reconstructed by the encoder $g_\phi(\cdot)$ and the decoder $h_\psi(\cdot)$.
    
    \subsection{Masked self-view modeling (MSM)}
    % The frames from the same video have causal relationships over time and \method learns this causality between frames through masked self-view modeling.
    % Similar to the previous masked modeling approaches~\cite{mae, videomae}, we adopt an asymmetric design 
    To learn the temporal dependency for fine-grained action recognition, we propose masked self-view modeling (MSM).
    We randomly sample a subset of frame tokens and remove the remaining ones.
    The sampled frame token embeddings from each view, $\bbXego^\text{MSM}$ and $\bbXexo^\text{MSM}$, are respectively mapped to the latent space that is used as a shared view-invariant embedding space, through the encoder $g_\phi(\cdot)$:
    \begin{equation}
    \begin{aligned}
        \bZego^\text{MSM} & = g_\phi(\bbXego^\text{MSM}) \in \mathbb{R}^{\Tegomsm \times d}, \\ \bZexo^\text{MSM} & = g_\phi(\bbXexo^\text{MSM}) \in \mathbb{R}^{\Texomsm \times d}.
    \end{aligned}
    \end{equation}
    The decoder $h_\psi(\cdot)$ reconstructs the original frame token embeddings (\ie, $\bbXego$ and $\bbXexo$) from the full set of frame tokens, consisting of frame tokens from the encoder (\ie, $\bZego'$ and $\bZexo'$) and learnable mask tokens~\cite{bert} filled into the position of removed tokens.
    We refer to this full set of frame tokens as ``mask-filled tokens".
    We add positional embeddings to all tokens to enable the decoder to take the dependency over time into account.
    To guarantee temporal dependency for mask prediction, we apply a causal mask~\cite{bert} to attention layers in the decoder so that the decoder reconstructs the $t$-th frame token only from the previous ($t-1$) tokens.
    
    The objective of MSM is defined by the mean squared error (MSE) between the reconstructed and original frame token embeddings:
    \begin{equation}
    \begin{aligned}
         \mathcal{L}_\text{MSM} =  \frac{1}{\Tego} || \bbXego & - \bYego^\text{MSM}||_2 + \frac{1}{\Texo} ||\bbXexo - \bYexo^\text{MSM}||_2, \\
        \text{where } & \bYego^\text{MSM} =  h_\psi (\bZego' || \texttt{[mask]}), \\
                        & \bYexo^\text{MSM} = h_\psi (\bZexo' || \texttt{[mask]}). \\
    \end{aligned}
    \end{equation}
    % where $\bYego^\text{MSM} = h_\psi (\bZego' || \texttt{[mask]})$ and $\bYexo^\text{MSM} = h_\psi (\bZexo' || \texttt{[mask]})$.
    We highlight that masking prediction is performed at the frame feature level, not the original frame image reconstruction.
    It can significantly reduce the computational training cost for the autoencoding process~\cite{wang2023masked}.
    Furthermore, it preserves strong spatial representation from the image encoder, while the encoder concentrates on learning temporal context with visual cues.
    \cref{fig:3b} illustrates the overall process of the MSM.
    % \begin{equation}
    %     \text{Attn} = \text{Softmax}(R + \log{(M + \epsilon)}),
    % \end{equation}
    
    \subsection{Masked cross-view modeling (MCM)}
    While MSM enables the model to capture the temporal dependency between frames within the same view video, masked cross-view modeling (MCM) aims to learn the view-invariant alignment between the different view videos.
    Similar to MSM, we first project the whole frame tokens (\ie, $\bbXego$ and $\bbXexo$) into the latent space, such that,
    \begin{equation}
        \bZego = g_\phi(\bbXego), \quad \bZexo = g_\phi(\bbXexo).
    \end{equation}
    Concurrently, we randomly remove a subset of frame tokens with a larger ratio than in MSM, and map to the latent space (see~\cref{fig:3c}):
    \begin{equation}
    \begin{aligned}
        \bZego^\text{MCM} & = g_\phi(\bbXego^\text{MCM}) \in \mathbb{R}^{\Tegomcm \times d}, \\
        \bZexo^\text{MCM} & = g_\phi(\bbXexo^\text{MCM}) \in \mathbb{R}^{\Texomcm \times d},
    \end{aligned}
    \end{equation}
    where $\bbXego^\text{MCM}$ and $\bbXexo^\text{MCM}$ are remaining frame tokens after masking.
    We note that the number of retention frames varies between ego and exo videos.
    In MCM, the decoder reconstructs the original frame token embeddings of one view video from the mask-filled tokens of its own view and the latents of another view:
    \begin{equation}
        \begin{aligned}
            \bYego^\text{MCM} = h_\psi(\bZego^\text{MCM} || \texttt{[mask]} || \bZexo), \\
            \bYexo^\text{MCM} = h_\psi(\bZexo^\text{MCM} || \texttt{[mask]} || \bZego). \\
        \end{aligned}
    \end{equation}
    The objective of MCM can be formulated by the MSE between the reconstructed and original frame token embeddings:
    \begin{equation}
        \mathcal{L}_\text{MCM} = \frac{1}{\Tego} ||\bbXego - \bYego^\text{MCM}||_2 + \frac{1}{\Texo} ||\bbXexo - \bYexo^\text{MCM}||_2.
    \end{equation}
    We force the cross-view generation as $\bZego^\text{MCM}$ and $\bZexo^\text{MCM}$ contain only a small subset (\eg 20\% in MCM vs.\ 60 \% in MSM) of the full tokens.
    Namely, the decoder is forced to reconstruct the original frame tokens from another view's tokens.
    This allows us to learn view-invariant latent space.

    \subsection{Joint training with MSM and MCM}
    The final objective of \method is the sum of the objectives of MSM and MCM:
    \begin{equation}
        \mathcal{L}_\method = \mathcal{L}_\text{MSM} + \mathcal{L}_\text{MCM}.
    \end{equation}
    While both loss terms serve a similar purpose (\ie, masked token embedding reconstruction), they operate differently based on the varying inputs to the decoder.
    The joint training aims to achieve both objectives simultaneously by enabling the encoder to map view-invariant features from ego-exo videos into a shared latent space.
    We notice that the forward-backward pass is separately performed by the loss, but the framework shares the same parameters.
% \subsection{Regularization}\label{sec:reg}

\section{Experiments}
\label{sec:4}

\begin{table*}[t]
    \centering
    \small
    \caption{Performance comparison with the state-of-the-art methods~\cite{aon, tcn, carl, gta, ae2} on the AE2 benchmark~\cite{ae2}. The benchmark consists of four sub-tasks: (A) Break Eggs, (B) Pour Milk, (C) Pour Liquid, and (D) Tennis Forehand.
    }\label{tab:1} % \vspace{-3pt}
    \setlength{\tabcolsep}{4pt}
        \begin{tabular}{clcccccccc}
        \toprule
        \multirow{2}[2]{*}{Task} & \multirow{2}[2]{*}{Method} & \multicolumn{3}{c}{Classification (F1 score)} & \multicolumn{3}{c}{Frame Retrieval (mAP@10)}& \multirow{2}[2]{*}{\makecell{ Phase\\ progression} } & \multirow{2}[2]{*}{\makecell{ Kendall's \\ $\tau$} } \\ 
        \cmidrule(lr){3-5} \cmidrule(lr){6-8}
       &  & Regular  & Ego2Exo & Exo2Ego    & Regular    & Ego2Exo   &   Exo2Ego   &       &                \\       
        \midrule
        \multirow{10}[2]{*}{(A)} & Random features &  19.18  &   18.93   &   19.45    &   47.13  &   41.74   &   37.19  &   -0.0572    &   0.0018   \\
        & ImageNet features &  50.24 & 21.48 &  32.25 &   50.49  &     33.09    &   37.80  &   -0.1446    &   0.0188  \\
        & CLIP features &  51.66 & 27.97 &  26.24 &   44.46  &     35.85    &   35.70  &   0.0402    &   0.0168  \\
        & ActorObserverNet~\cite{aon} &  36.14 & 36.40 &  31.00 &   50.47    &   42.70    &   41.29    &   -0.0517   &   0.0024    \\
        & TCN~\cite{tcn} (single-view) &  56.90 & 18.60 &  35.61 &   53.42  &     32.63    &   34.91  &   0.0051    &   0.1206   \\
        & TCN~\cite{tcn} (multi-view) &  59.91 & 48.65 &  56.91 &   58.83  &     47.04    &   52.68  &   0.2669    &   0.2886   \\
        & CARL~\cite{carl} &  43.43 & 28.35 &  29.22 &   46.04  &     37.38    &   39.94  &   -0.0837   &   -0.0091   \\
        & TCC~\cite{tcc} &  59.84 & 54.17 &  52.28 &   58.75  &     61.11    &   62.03  &   0.2880   &   0.5191   \\
        & GTA~\cite{gta} &  56.86 & 52.33 &  58.35 &   61.55  &     56.25    &   53.93  &   0.3462   &   0.4626   \\
        & AE2~\cite{ae2} &  {66.23} & {57.41} &  \textbf{71.72} &   {65.85}  &     {64.59}    &   {62.15}  &   {0.5109}   &   {0.6316}   \\
        \cmidrule(lr){2-10}
        & \method  &  \textbf{74.30} & \textbf{75.01} &  {71.28} &   \textbf{67.17}  &     \textbf{70.65}    &   \textbf{69.02}  &   \textbf{0.8533}   &   \textbf{0.9451}   \\
        \midrule
        \multirow{8}[2]{*}{(B)} & Random features &  36.84  &   33.96   &   41.97    &   52.48  &   50.56   &   51.98  &   -0.0477    &   0.0050   \\
        & ImageNet features &  41.59 & 39.93 &  45.52 &   54.09  &     27.31    &   43.21  &   -2.6681    &   0.0115  \\
        & CLIP features &  43.24 & 49.21 &  30.94 &   52.16  &     46.39    &   40.34  &   -4.0754    &   0.0046  \\
        & TCN~\cite{tcn} (single-view) &  47.39 & 43.44 &  42.28 &   57.00    &   46.48    &   47.20    &   -0.3238   &   -0.0197    \\
        & CARL~\cite{carl} &  48.79 & 52.41 &  43.01 &   55.01  &     52.99    &   51.51  &   -0.1639   &   0.0443   \\
        & TCC~\cite{tcc} &  77.91 & 72.29 &  81.07 &   80.97  &     75.30    &   80.27  &   0.6665   &   0.7614   \\
        & GTA~\cite{gta} &  81.11 & 74.94 &  81.51 &   80.12  &     72.78    &   75.40  &   0.7086   &   0.8022   \\
        & AE2~\cite{ae2} &  85.17 & 84.73 &  82.77 &   84.90  &     78.48    &   {83.41}  &   0.7634   &   0.9062   \\
        % \textsc{DeCap}~\citep{decap} &   &   &   \checkmark    & CLIP & CLIP & 50.6 &  -    \\
        \cmidrule(lr){2-10}
        % \textbf{Ours}   &    \checkmark     &   &    &  CLIP &    CLIP   &  \bf 58.3  &  \bf 81.7    \\
        % \textbf{Ours}   &    \checkmark     &   &    &  CLIP &    GPT-2   &   56.8  &   79.1    \\
        % \textbf{Ours}   &    \checkmark     &   &    &  BEiT &    GPT-2   &   53.2  &   71.7    \\
        % \textbf{Ours}   &    & \checkmark  &   \checkmark    &  CLIP &    CLIP   &  37.1  &  69.5    \\
        % \textbf{Ours}   &    & \checkmark  &   \checkmark    &  BEiT &    GPT-2   &  -  &  -    \\
        & \method  &  \textbf{86.46} & \textbf{85.09} &  \textbf{86.61} &   \textbf{89.42}  &     \textbf{87.73}    &   \textbf{85.06}  &   \textbf{0.8992}   &   \textbf{0.9466}   \\
        \midrule
        \multirow{8}[2]{*}{(C)} & Random features &  45.26  &   47.45   &   44.33    &   49.83  &   55.44   &   55.75  &   -0.1303    &   -0.0072   \\
        & ImageNet features &  53.13 & 22.44 &  44.61 &   51.49  &     52.17    &   30.44  &   -1.6329    &   -0.0053  \\
        & CLIP features &  60.60 & 36.97 &  48.43 &   43.63  &     47.58    &   37.02  &   -0.3139    &   -0.0048  \\
        % Flamingo &  10.2B & 1.8B &  \textbf{52.0} &   -  & -    &   -   &   -     & \multicolumn{3}{c}{Incapable}  \\
        & TCN~\cite{tcn} (single-view) &  54.02 & 32.77 &  51.24 &   48.83  &     55.28    &   31.15  &   -0.5283    &   0.0103   \\
        & CARL~\cite{carl} &  56.98 & 47.46 &  52.68 &   55.29  &     59.37    &   36.80  &   -0.1176   &   0.0085   \\
        & TCC~\cite{tcc} &  52.53 & 43.85 &  42.86 &   62.33  &     56.08    &   57.89  &   0.1163   &   0.1103   \\
        & GTA~\cite{gta} &  56.92 & 42.97 &  59.96 &   62.79  &     58.52    &   53.32  &   -0.2370   &   0.1005   \\
        & AE2~\cite{ae2} &  66.56 & 57.15 &  65.60 &   {65.54}  &     65.79    &   57.35  &   0.1380   &   0.0934   \\
        % \textsc{DeCap}~\citep{decap} &   &   &   \checkmark    & CLIP & CLIP & 50.6 &  -    \\
        \cmidrule(lr){2-10}
        % \textbf{Ours}   &    \checkmark     &   &    &  CLIP &    CLIP   &  \bf 58.3  &  \bf 81.7    \\
        % \textbf{Ours}   &    \checkmark     &   &    &  CLIP &    GPT-2   &   56.8  &   79.1    \\
        % \textbf{Ours}   &    \checkmark     &   &    &  BEiT &    GPT-2   &   53.2  &   71.7    \\
        % \textbf{Ours}   &    & \checkmark  &   \checkmark    &  CLIP &    CLIP   &  37.1  &  69.5    \\
        % \textbf{Ours}   &    & \checkmark  &   \checkmark    &  BEiT &    GPT-2   &  -  &  -    \\
        & \method  &  \textbf{79.48} & \textbf{71.83} &  \textbf{76.23} &   \textbf{71.06}  &     \textbf{75.03}    &   \textbf{70.03}  &   \textbf{0.4483}   &   \textbf{0.3052}   \\
        \midrule
        \multirow{8}[2]{*}{(D)} & Random Features &  30.31  &   33.42   &   28.10    &   66.47  &   58.98   &   59.87  &   -0.0425    &   0.0177   \\
        & ImageNet Features &  69.15 & 42.03 &  58.61 &   76.96  &     66.90    &   60.31  &   -0.4143    &   0.0734  \\
        & CLIP Features &  67.81 & 43.41 &  44.22 &   74.54  &     59.57    &   52.02  &   -0.4996    &   0.0618  \\
        % Flamingo &  10.2B & 1.8B &  \textbf{52.0} &   -  & -    &   -   &   -     & \multicolumn{3}{c}{Incapable}  \\
        & TCN~\cite{tcn} (single-view) &  68.87 & 48.86 &  36.48 &   73.76  & 55.08 & 56.65 & -0.0602    &   0.0737   \\
        & CARL~\cite{carl} &  59.69 & 35.19 &  47.83 &   69.43  & 54.83 & 63.19 & -0.1310 & 0.0542   \\
        & TCC~\cite{tcc} &  78.41 & 53.29 & 32.87 & 80.24 & 55.84 & 63.19 & 0.2155 & 0.1040   \\
        & GTA~\cite{gta} &  83.63 & 82.91 & 81.80 & 85.20 & 78.00 & 79.14 & 0.4691 & 0.4901   \\
        & AE2~\cite{ae2} &  85.87 & 84.71 & 85.68 & 86.83 & 81.46 & 82.07 & 0.5060 & 0.6171   \\
        % \textsc{DeCap}~\citep{decap} &   &   &   \checkmark    & CLIP & CLIP & 50.6 &  -    \\
        \cmidrule(lr){2-10}
        % \textbf{Ours}   &    \checkmark     &   &    &  CLIP &    CLIP   &  \bf 58.3  &  \bf 81.7    \\
        % \textbf{Ours}   &    \checkmark     &   &    &  CLIP &    GPT-2   &   56.8  &   79.1    \\
        % \textbf{Ours}   &    \checkmark     &   &    &  BEiT &    GPT-2   &   53.2  &   71.7    \\
        % \textbf{Ours}   &    & \checkmark  &   \checkmark    &  CLIP &    CLIP   &  37.1  &  69.5    \\
        % \textbf{Ours}   &    & \checkmark  &   \checkmark    &  BEiT &    GPT-2   &  -  &  -    \\
        & \method  &  \textbf{89.12} & \textbf{94.47} &  \textbf{85.73} &   \textbf{90.61}  &     \textbf{88.34}    &   \textbf{88.94}  &   \textbf{0.7881}   &   \textbf{0.7852}   \\
        \bottomrule
        \end{tabular}
        % \vspace{-5pt}
    \end{table*}
    \subsection{Evaluation setup}
    % \paragraph{Evaluation.}
    The essential capabilities of fine-grained, view-invariant video representations are 1) fine-grained temporal understanding of the given action and 2) alignment across cross-view videos. 
    Following previous work~\cite{ae2}, we evaluate \method on four datasets in the ego-exo benchmark~\cite{ae2}, including \textit{Break Eggs}, \textit{Pour Milk}, \textit{Pour Liquid}, and \textit{Tennis Forehand}.
    Each dataset consists of four tasks as follows:
    \begin{itemize}
        \item {\textbf{Action phase classification} aims to predict an atomic action phase label corresponding to a given frame.
        For example, `Break Eggs' dataset contains four action phases following the key events, such as `hit egg', `visible crack on the eggshell', `egg contents released into bowl.'
        Following \cite{ae2}, we train an SVM classifier with the embeddings from the training set and evaluate the F1 score on the test set.
        We evaluate the performance in two settings: the regular setting, where the classifier is trained using embeddings from both views, and the cross-view setting, where training and testing data are sourced from different view videos.}
        
        \item {\textbf{Frame retrieval} shares the same context with classification in evaluation, but does not require additional training. 
        We evaluate the performance with Nearest Neighbor search (NN) to retrieve frames and report mean average precision (mAP)@10 performance in the regular and cross-view settings.}
        
        \item {\textbf{Phase progression} evaluates how accurately the embeddings capture the progress of an action over time by predicting the phase progression values defined as the time-stamp difference between each frame and key events, normalized by the total number of frames in the video~\cite{tcc}.
        We train a linear regressor with the embeddings and evaluate the average R-square. }
        
        \item {\textbf{Kendall's $\boldsymbol\tau$} measures how well-aligned two given sequences are in time. 
        We first sample a pair of frames in one video and retrieve the corresponding nearest frames from another video.
        Kendall's $\tau$ measures concordant, which represents whether the order of the retrieved frame pair matches the order of the frame pair in the first video, over every pair of frames in a pair of videos.}
    \end{itemize}
    Each dataset contains a varying number of subjects, and each subject is captured from different equipment.
    A detailed description of the evaluation benchmark is demonstrated in \cref{sec:dataset}.

    In addition, we evaluate video-level action recognition performance on the Charades-Ego dataset~\cite{charades-ego}, which contains 4k paired ego-exo videos, using the linear probing protocol.
    Specifically, our model is trained on the Charades-Ego training set, and video-level embeddings are averaged by frame-level embeddings.
    We train a linear classifier with video embeddings and report mAP scores on the test set.

    % \paragraph{Model architecture.} 
    \subsection{Implementation details}
    We employ CLIP pretrained ViT-B/16~\cite{vit} as the image encoder $f_\theta(\cdot)$, and freeze it.
    We take the last hidden states of the image encoder except for a class token.
    The encoder and decoder for masked modeling have 12 and 4 Transformer blocks, respectively, with 256-dimensional embedding space.
    We learn frame embedding encoder $g_\phi(\cdot)$ and the decoder $h_\psi(\cdot)$ on the training set, while $g_\phi(\cdot)$ is frozen during the training of SVM classifier and the linear regression for evaluation.
    The decoder is discarded after training.
    Note that the encoder and decoder have only 9.7M and 2.6M parameters respectively.
    We empirically set the token selection ratio to 0.3 and the masking ratio in the self-view prediction and cross-view prediction to 0.4 and 0.8, respectively.
    % The ablation studies for the token selection ratio and the masking ratio are provided in \cref{sec:additional}.
    % \method is first trained on the training set, while keeping frozen the image encoder $f_{\theta}(\cdot)$ and frame embedding encoder $g_\phi(\cdot)$ when the SVM classifier and the linear regressor are trained at evaluation.
    % The decoder $h_\psi(\cdot)$ is discarded after training.
    For the Break Eggs, Pour Milk, and Pour Liquid datasets, we randomly sample 32 frames to cover the whole video for training and use all frames for evaluation.
    For the Tennis Forehand dataset, we use randomly sampled 20 frames for training as they have shorter video lengths.
    For Charades-Ego, we randomly sample 32 frames for training.
    While ego-exo videos are synchronized, we do not use the temporal correspondence between videos for training.
    
    \paragrapht{Baselines.}
    We compare the performance of \method against a variety of baselines with different characteristics:
    (1) Randomly initialized feature (\textbf{Random features}), \textbf{ImageNet features} from the ResNet-50~\cite{resnet} pretrained on ImageNet~\cite{imagenet}, and \textbf{CLIP features} from the CLIP ViT-B/16 pretrained on LAION-400M~\cite{laion-400m} which is used as the frame feature in \method.
    (2) Self-supervised view-invariant video representation learning (\textbf{TCN}~\cite{tcn}, \textbf{ActorObserverNet}~\cite{aon}).
    (3) Fine-grained video representation learning (\textbf{CARL}~\cite{carl}, \textbf{TCC}~\cite{tcc}, \textbf{GTA}~\cite{gta}).
    (4) Fine-grained view-invariant video representation learning (\textbf{AE2}~\cite{ae2}) that is directly comparable to our \method.

    \subsection{Results}

       \begin{figure}[t]
      \centering
        \begin{subfigure}{0.9\linewidth}
        \centering
        {\includegraphics[width=0.98\linewidth]{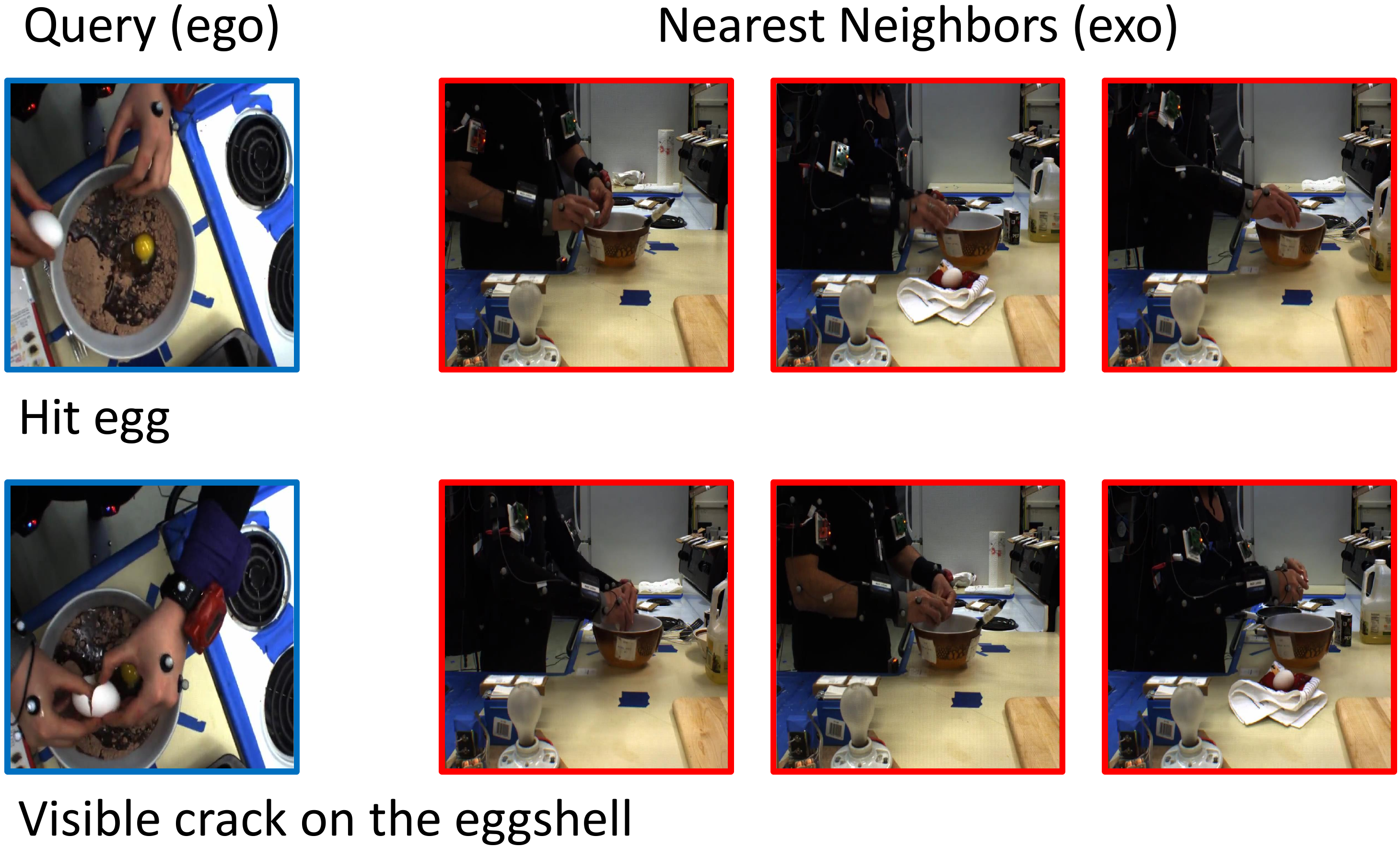}}
        \caption{Ego2Exo frame retrieval on Break Eggs}\label{fig:quala}
       \end{subfigure}  \\
       \begin{subfigure}{0.9\linewidth}
        \centering
        {\includegraphics[width=0.98\linewidth]{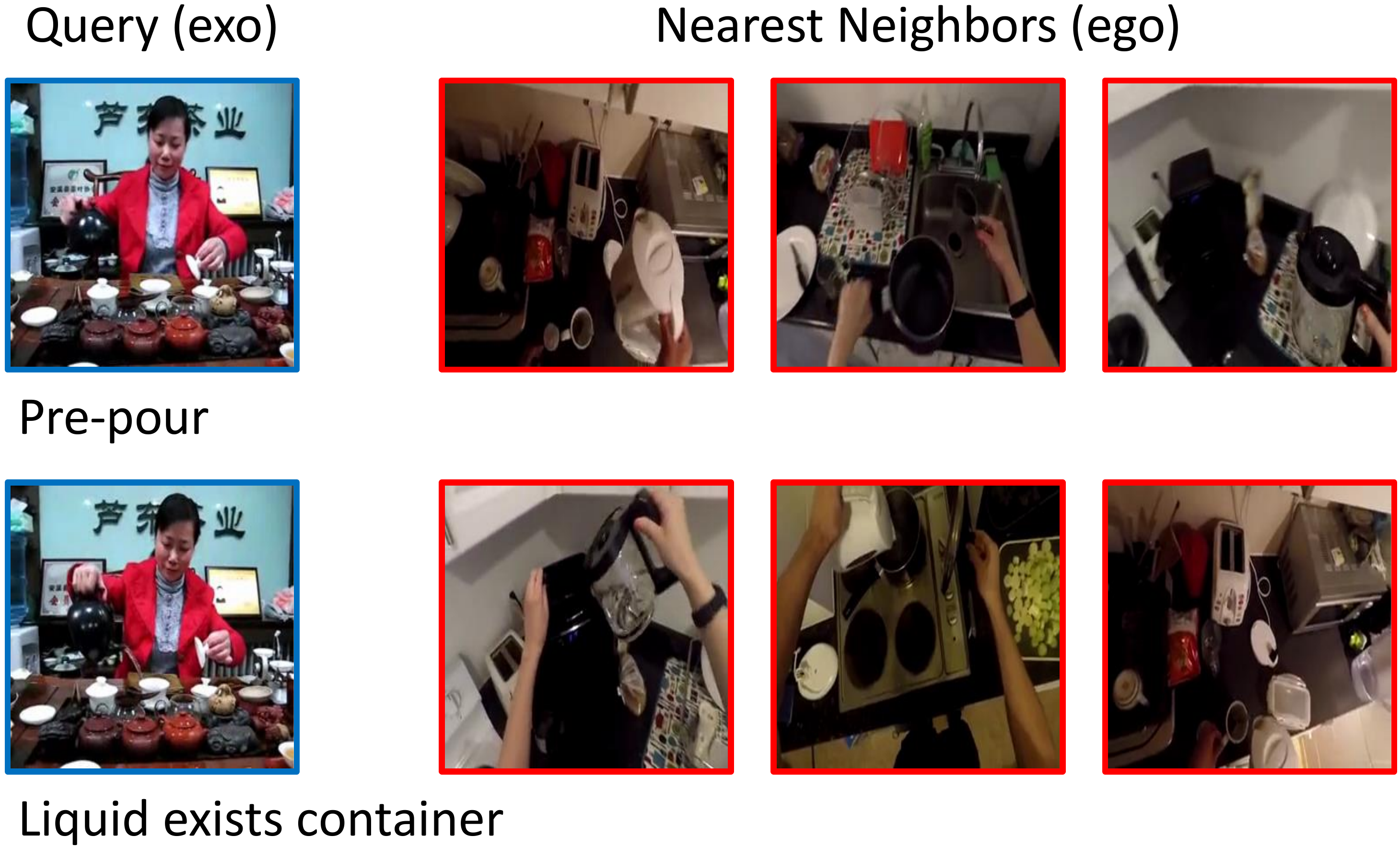}}
        \caption{Exo2Ego frame retrieval on Pour Liquid}\label{fig:qualb}
       \end{subfigure}%\vspace{-3pt}
      \caption{Results of frame retrieval from \textit{Break Eggs} and \textit{Pour Liquid}. We retrieve the nearest neighbor frames (red box) corresponding to the given query frame (blue box).}%\vspace{-7pt}
      \label{fig:qual}
    \end{figure}

\begin{figure*}[t]
  \centering
    \begin{subfigure}{0.48\linewidth}
    \centering
    {\includegraphics[width=1.0\linewidth]{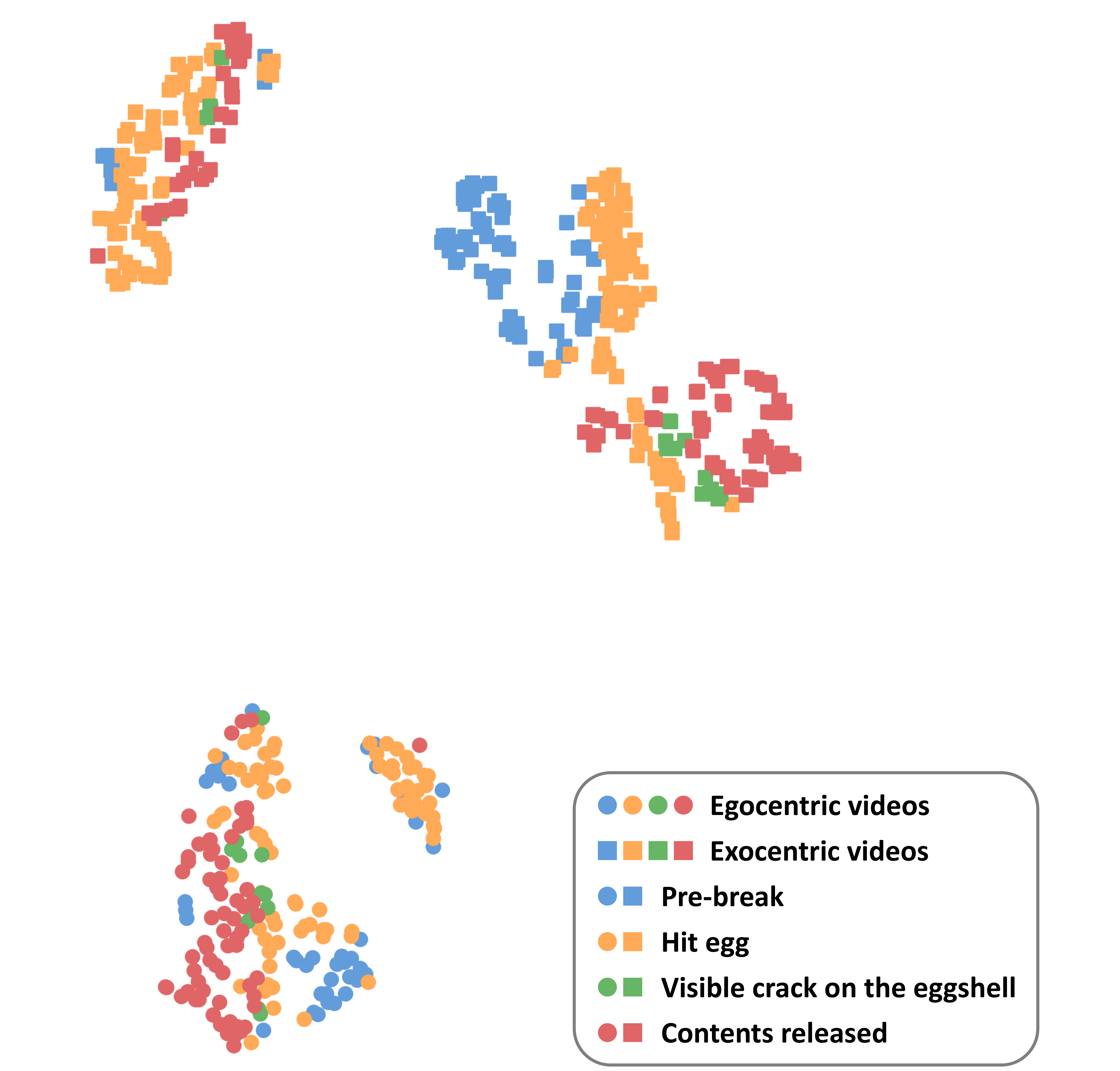}}
    \caption{Frame embeddings before training}\label{fig:visualizationa}
   \end{subfigure} \hfill % \\  \vspace{5pt}
   \begin{subfigure}{0.48\linewidth}
    \centering
    {\includegraphics[width=1.0\linewidth]{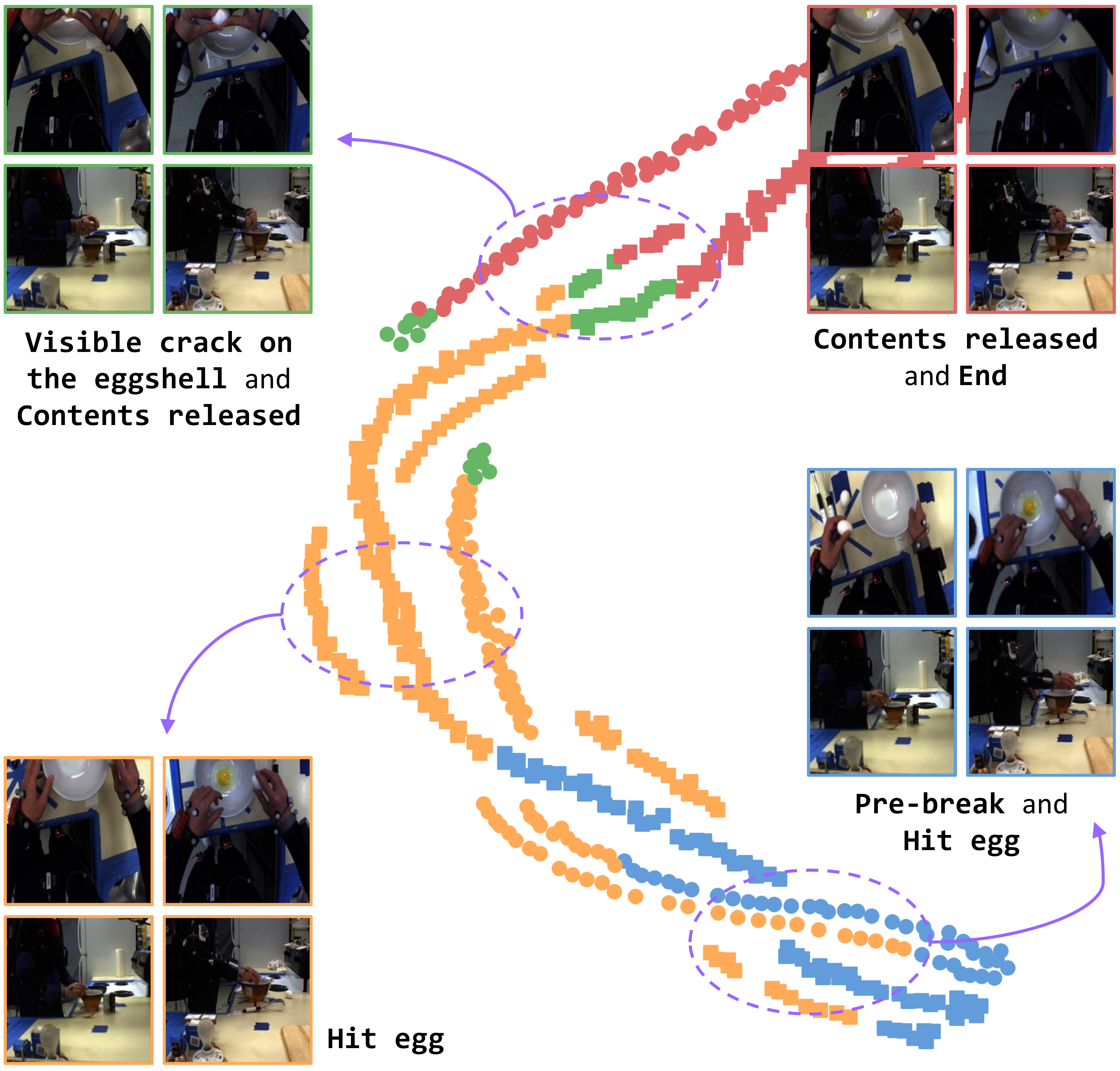}}
    \caption{Frame embeddings of \method}\label{fig:visualizationb}
   \end{subfigure}
  \caption{tSNE visualization of frame embeddings (a) before training and (b) trained with \method. We sample two ego and two exo videos from \textit{Break Eggs}.}
  \label{fig:visualization}
\end{figure*}

    \paragrapht{Quantitative comparisons on AE2.}
    In \cref{tab:1}, we present the performance of each task across four ego-exo datasets.
    The comparison between ImageNet and CLIP features reveals that the differences between the encoder architecture (\ie, CNNs~\cite{resnet} vs. transformers~\cite{vit}), and the scale of the pretraining dataset (\ie, 1.3M~\cite{imagenet} vs. 400M~\cite{laion-400m}) have minimal impact on fine-grained, view-invariant video understanding.
    Notably, both features struggle to capture the progress of an action over time especially in the cross-view.
    Consequently, they show poor performance in phase progression and Kendall's $\tau$ tasks even than random features.
    Meanwhile, \method significantly outperforms existing SoTAs~\cite{gta,ae2}. 
    The comparison between baselines and \method demonstrates the powerful representative capability of \method.
    Especially, our \method outperforms others in in-the-wild scenarios as shown in the `Pour Liquid' and `Tennis Forehand' datasets.
    It illustrates that our masked modeling is more effective self-supervision than previous works~\cite{tcn, aon} for real-world videos.
    In \cref{sec:additional}, we provide additional results with CLIP pretrained ViT-L/14~\cite{vit}, and also provide the F1 scores for few-shot classification, mAP@5, and maP@15 for frame retrieval.

    \paragrapht{Qualitative results on AE2.}
    In \cref{fig:qual}, we depict examples for cross-view frame retrieval from the Break Eggs and Pour Liquid datasets.
    We retrieve the frames (red box) in all test frames from the query frame (blue box) using NN search.
    The query frame and retrieved frames are obtained from different views.
    The results show that the query and retrieved frames are contextually connected through the action state despite the significant visual discrepancy between the ego-exo views.
    In \cref{fig:quala}, for example, egocentric representations from \method successfully distinguish `hit egg' and `visible crack on the eggshell' phases to retrieve corresponding frames in exocentric videos. 
    More qualitative results are illustrated in \cref{sec:retrieval_supp}.

    \paragrapht{tSNE visualization on AE2.}
    \cref{fig:visualization} illustrates tSNE visualization of frame embeddings (a) before training and (b) trained with \method.
    We sample four videos (two ego and two exo videos) from the Break Eggs dataset, and distinguish views and action phases by shapes of the point (circle for ego, and square for exo) and colors, respectively.
    Before training, we can observe a clear gap between the frame embeddings (from the ViT-B/16) according to the viewpoint and video ID, while the embeddings clumped together regardless of the action phases, as shown in \cref{fig:visualizationa}.
    We argue that pretrained image encoders are hard to capture discriminative features for the action phase due to the large visual similarity between frames.
    Surprisingly, the frame representations obtained from \method form trajectories according to the action phase, regardless of the viewpoint, as shown in \cref{fig:visualizationb}.
    The result demonstrates that \method effectively learns temporal dynamics and accomplishes view invariance.

    \paragrapht{Results on Charades-Ego.}
    We compare the performance of Random features, CLIP features from the CLIP ViT-B/16, and \method with the CLIP ViT-B/16 on Charades-Ego.
    We measure mAP scores using linear probing with three evaluation settings, as shown in \cref{tab:charades}: `Regular,' which utilizes both ego and exo videos for training; `Ego2Exo,' where the classifier is trained on ego videos and tested on exo videos; and `Exo2Ego,' which performs the reverse.
    \method significantly improves mAP scores by 18.1\%, 18.2\%, and 16.6\% in each setting, demonstrating its effectiveness even in the video-level action understanding.

\begin{table}[t]
    \centering
    \small
    \caption{Performance comparison on Charades-Ego.
    }\label{tab:charades} % \vspace{-5pt}
    \setlength{\tabcolsep}{4pt}
        \begin{tabular}{lccc}
        \toprule
        {Method}     & Regular  &   Ego2Exo  &  Exo2Ego  \\
        \midrule
        Random features  &   7.2 &   7.1   &   7.2   \\
        % ResNet-152 (FT)  &   25.7   &   -   &   -   \\
        CLIP features (ViT-B/16)    & 13.7    &   8.3   &   10.7  \\
        \midrule
        % \method     &   \textbf{73.53}   &   \textbf{68.95}   &   \textbf{0.8533}  &   \textbf{0.9451}  \\
        \method (ViT-B/16)   &   \textbf{31.8}   &   \textbf{26.5}   &  \textbf{27.3}   \\
        % % \multicolumn{5}{l}{\hspace{-5pt}\textit{Effectiveness of Each Path}} \\ 
        % \, -- Token selection   &  71.84  &  67.55 & 0.8224  &  0.9016   \\
        % \, -- Causal mask   &  72.29  &  68.10 & 0.6420  &  0.7091   \\
        % \, -- MSM   &  62.12  &  60.34 & 0.4362  &  0.4906   \\
        % \, -- MCM   &  57.43  &  55.83 & 0.6724  &  0.7086   \\
        \bottomrule
        \end{tabular}
        % \vspace{-5pt}
    \end{table}

\begin{table}[t]
    \centering
    \small
    \caption{Performance with respect to variants of the components in \method. We report the performance evaluated on \textit{Break Eggs}.
    }\label{tab:ablation} % \vspace{-3pt}
    \setlength{\tabcolsep}{4pt}
        \begin{tabular}{lcccc}
        \toprule
        {Method}     & F1 score  &  mAP@10  &  Progr.   & $\tau$   \\
        \midrule
        \method     &   \textbf{73.53}   &   \textbf{68.95}   &   \textbf{0.8533}  &   \textbf{0.9451}  \\
        % \multicolumn{5}{l}{\hspace{-5pt}\textit{Effectiveness of Each Path}} \\ 
        \, -- Token selection   &  71.84  &  67.55 & 0.8224  &  0.9016   \\
        \, -- Causal mask   &  72.29  &  68.10 & 0.6420  &  0.7091   \\
        \, -- MSM   &  62.12  &  60.34 & 0.4362  &  0.4906   \\
        \, -- MCM   &  57.43  &  55.83 & 0.6724  &  0.7086   \\
        \bottomrule
        \end{tabular}%\vspace{-3pt}
    \end{table}

    \subsection{Component analysis}

    %    \begin{figure}[t]
    %   \centering
    %     \begin{subfigure}{0.9\linewidth}
    %     \centering
    %     {\includegraphics[width=0.98\linewidth]{fig/fig-maska.pdf}}
    %     \caption{Masked token visualization on Break Eggs}\label{fig:maska}
    %    \end{subfigure}  \\
    %    \begin{subfigure}{0.9\linewidth}
    %     \centering
    %     {\includegraphics[width=0.98\linewidth]{fig/fig-maskb.pdf}}
    %     \caption{Masked token visualization on Tennis Forehand}\label{fig:maskb}
    %    \end{subfigure}\vspace{-3pt}
    %   \caption{Visualization of selected tokens from the Break Eggs and Tennis Forehand datasets. The selection ratio is 0.3.  Our simple selective token merging approach well picks the action-related regions. The excluded patches are black.}
    %   \label{fig:mask}
    % \end{figure}
    
    \paragrapht{Token selection.}
    We select $K$ tokens based on the difference between token embeddings from consecutive frames to encode the action-related spatial information.
    Since fine-grained action recognition requires detecting subtle yet critical movements, it is important to utilize selective features effectively without noisy redundant information. 
    % In practice, action-related tokens can be selected by simply comparing the difference of token embeddings, as shown in \cref{fig:mask}.
    To verify the effectiveness of token selection, we evaluate the performance without token selection as shown in \cref{tab:ablation} (\method $-$Token selection).
    The result shows that selective token merging improves the performance in all metrics.
    Ablation studies and visualization corresponding to the token selection ratio are provided in \cref{sec:ablation_append}.

    \paragrapht{Causal mask in MSM.}
    We apply the causal mask in the decoder during MSM to enable the model to learn the temporal dependency between frames.
    \cref{tab:ablation} (\method $-$Causal mask) demonstrates the impact of the causal mask. 
    Especially, the performance is notably degraded in both phase progression and Kendall's $\tau$, showing the significance of apprehending causality to downstream tasks.
    % We evaluate the impact of the causal mask 

    \paragrapht{Masked modeling.}
    % We analyze the effectiveness of masked self-view and cross-view modeling.
    \cref{tab:ablation} demonstrates the effectiveness of the choice of masked modeling methods.
    \method $-$MSM shows that the performance significantly drops across all tasks, especially in phase progression and Kendall's $\tau$ that are highly related to the causality between frames.
    The results demonstrate the effectiveness of MSM in learning the temporal dependency for fine-grained action recognition.
    Meanwhile, the ablation study for MCM (\method $-$MCM) shows significant performance drops in the cross-view classification and retrieval tasks.
    The results demonstrate that MCM and MSM complement each other from different perspectives.
    Both masked modelings enable robust view-invariant learning across egocentric and exocentric videos, thereby effects for fine-grained video understanding.
    % Consequently, 
    We provide the effectiveness of representations corresponding to the masking ratio in MSM and MCM in \cref{sec:ablation_append}.
    % \paragraph{Masking ratio in MSM.}

    % \paragraph{Masking ratio in MCM.}

\section{Conclusion}
\label{sec:5}

We propose \method to learn view-invariant video representations with masked ego-exo modeling.
During training, our method simultaneously predicts self- and cross-view feature masking tokens in a self-supervised manner with unpaired and asynchronous egocentric and exocentric videos.
While masked self-view modeling boosts the model to capture fine-grained temporal dependency in actions, masked cross-view modeling accelerates cross predictions by view-invariant video representations.
Moreover, our selective token merging approach is simple, yet can be an effective alternative for off-the-shelf human-object interaction detectors or additional temporal dynamic signals, without the additional burden of computational cost.
\method achieves state-of-the-art performance with significantly larger margins than existing methods for cross-view invariant learning in four downstream video tasks.
% We remain such tasks in our future work.

\paragrapht{Acknowledgement.}
This work was supported by the Institute of Information \& communications Technology Planning \& Evaluation (ITP) grant funded by the Korea government (MSIT) (No.~RS-2022-00155966, Artificial Intelligence Convergence Innovation Human Resources Development (Ewha Womans University)).
{
    \small
    \bibliographystyle{ieeenat_fullname}
    \bibliography{main}
}

\clearpage
\newpage
\appendix
\setcounter{table}{0}
\renewcommand{\thetable}{A\arabic{table}}
\setcounter{figure}{0}
\renewcommand{\thefigure}{A\arabic{figure}}

\section*{Appendix}

    In this document, we provide more concrete details of the AE2 benchmark~\cite{ae2} in \cref{sec:dataset}, additional experimental results in \cref{sec:additional}, including the results using different backbones (CLIP pretrained ViT-L/14~\cite{vit} and ResNet-50~\cite{resnet}), few-shot classification and frame retrieval performance, ablation studies for each hyper-parameter, and analysis for the failure case.
    Finally, we present the broader impact of our \method in \cref{sec:impact}.
    % we present additional qualitative results to analyze the effectiveness of our \method in \cref{sec:qual}.

\section{Benchmark Details}\label{sec:dataset}
    In this section, we provide a detailed explanation of AE2 benchmark~\cite{ae2} and the evaluation details of four downstream tasks, including action phase classification, frame retrieval, phase progression, and Kendall's $\tau$.
    \subsection{Datasets}
    The AE2 benchmark~\cite{ae2} contains four datasets: (1) Break Eggs; (2) Pour Mild; (3) Pour Liquid; and (4) Tennis Forehand.
    The summary of each dataset is shown in \cref{tab:dataset}.
    \begin{itemize}
        \item {\textbf{Break Eggs} sampled from the CMU-MMAC dataset~\cite{cmu-mmac} contains 5 different cooking recipes (brownies, pizza, sandwiches, salad, and scrambled eggs) captured by 43 users.
        While the ego and exo videos are strictly synchronized (i.e., capturing the same scene), we do not use the correspondence between videos for training.
        }
        \item {\textbf{Pour Milk} sampled from the H2O dataset~\cite{h2o} contains the scene of 10 users interacting with a milk carton using their hands.
        The dataset provides one egocentric video and four exocentric static videos for each scene.
        Some ego and exo video pairs are synchronized and the rest are asynchronous.
        }
        \item {\textbf{Pour Liquid} assumes a more challenging scenario as the ego and exo videos are sampled from different datasets.
        Therefore, those videos are fully asynchronous and captured from different environments.
        The ego videos consist of the ``pour water" class in EPIC-Kitchens~\cite{damen2018scaling} and the exo videos are the ``pour" category in HMDB51~\cite{hmdb51}.
        }
        \item {\textbf{Tennis Forehand} includes outdoor activity videos.
        The exocentric videos of the tennis forehand action are sampled from the Penn Action~\cite{pennaction} dataset and the egocentric videos are collected from 12 players using the Go Pro HERO8 camera.
        The videos are asynchronous, covering real-world scenarios.
        }
    \end{itemize}

    \begin{table*}[t]
    \centering
    \small
    \caption{Performance with respect to variants of the components in \method. We report the performance evaluated on the Break Eggs dataset.
    }\label{tab:dataset}
    \setlength{\tabcolsep}{4pt}
        \begin{tabular}{lcccccccc}
        \toprule
        \multirow{2}[2]{*}{Dataset} & \multicolumn{2}{c}{\# Train} & \multicolumn{2}{c}{\# Val} & \multicolumn{2}{c}{\# Test} &  \multirow{2}[2]{*}{\makecell{ Fixed \\ exo-view} } & \multirow{2}[2]{*}{\makecell{ Sync. \\ ego-exo }} \\
        \cmidrule(lr){2-3} \cmidrule(lr){4-5} \cmidrule(lr){6-7} 
        &   Ego & Exo & Ego & Exo & Ego & Exo & & \\
        \midrule
        (A) Break Eggs     & 61 & 57 & 5 & 5 & 10 & 10 & \cmark & \cmark  \\
        (B) Pour Milk     & 29 & 48 & 4 & 8 & 7 & 16 & \cmark & \xmark  \\
        (C) Pour Liquid     & 70 & 67 & 10 & 9 & 19 & 18 & \xmark & \xmark  \\
        (D) Tennis Forehand     & 94 & 79 & 25 & 24 & 50 & 50 & \xmark & \xmark  \\
        \bottomrule
        \end{tabular}
    \end{table*}

    \subsection{Downstream tasks}
    \begin{itemize}
        \item {\textbf{Action phase classification} aims to predict an atomic action phase label corresponding to a given frame.
        The Break Eggs dataset contains four action phases between `start', `hit egg', `visible crack on the eggshell', `egg contents released', and `end.'
        The Pour Milk and Pour Liquid datasets contain three phases between `start', `liquid exits container', `pouring complete', and `end.'
        The Tennis Forehand dataset has only two phases between `start', `racket touches ball', and `end.'
        In this document, we additionally provide a few-shot classification performance to validate the robustness of \method.
        }
        \item {\textbf{Frame retrieval} selects frames corresponding to a given frame using the NN search.
        We evaluate this task with mean average precision (mAP)@K (K=5,10,15) in the regular and cross-view settings.
        }
        \item {\textbf{Phase progression} quantifies how effectively the learned representations imply the progression of an action. 
        The progression value within each phase is defined as the normalized temporal difference between the timestamp of a given frame and those of key events, scaled by the total number of frames in the video. 
        A linear regressor is then employed to predict the phase progression values from the embeddings, where our encoders are frozen. 
        The performance is evaluated using the average $R$-squared value as follows:
        \begin{align*}
            R^2 = 1 - \frac{\sum_{t=1}^T (y_t - \hat{y}_t)^2}{\sum_{t=1}^T (y_t - \bar{y})^2},
        \end{align*}
        where $y_t$ is the ground truth phase progress value, $\bar{y}$ is the average value of all $y_t$, and $\hat{y}_t$ is the prediction from the linear regressor.
        The maximum value of $R^2$ is 1.
        }
        \item {\textbf{Kendall's $\boldsymbol\tau$} assesses the temporal alignment between two sequences by comparing the order of frames. 
        Specifically, we first sample a pair of frames from one video, $(u_i, u_j)$, and retrieve their nearest corresponding frames in the other video, $(v_p, v_q)$. 
        A set of frame indices $(i, j, p, q)$ is treated as `matched' if the temporal order of $u_i$ and $u_j$ and that of $v_p$ and $v_q$ are the same.
        Kendall's $\tau$ is then computed by,
        \begin{align*}
            \tau = \frac{\# \text{matched pairs} - \# \text{not matched pairs}}{\# \text{all possible pairs}}.
        \end{align*}
        A value of $1$ means the frame representations are perfectly aligned while $-1$ indicates the representations are aligned in the reverse order.
        }
    \end{itemize}
\section{Additional Experiments}\label{sec:additional}
\begin{table*}[t]
    \centering
    \small
    \caption{Performance comparison with various frame encoders on the AE2 benchmark~\cite{ae2}. The benchmark consists of four sub-tasks: (A) Break Eggs, (B) Pour Milk, (C) Pour Liquid, and (D) Tennis Forehand. The top results are highlighted in \textbf{bold} and the second-best results are \underline{underlined}.
    }\label{tab:vitL}\vspace{-3pt}
    \setlength{\tabcolsep}{4pt}
        \begin{tabular}{clcccccccc}
        \toprule
        \multirow{2}[2]{*}{Task} & \multirow{2}[2]{*}{Method} & \multicolumn{3}{c}{Classification (F1 score)} & \multicolumn{3}{c}{Frame Retrieval (mAP@10)}& \multirow{2}[2]{*}{\makecell{ Phase\\ progression} } & \multirow{2}[2]{*}{\makecell{ Kendall's \\ $\tau$} } \\ 
        \cmidrule(lr){3-5} \cmidrule(lr){6-8}
       &  & Regular  & Ego2Exo & Exo2Ego    & Regular    & Ego2Exo   &   Exo2Ego   &       &                \\       
        \midrule
        \multirow{10}[2]{*}{(A)} & Random features &  19.18  &   18.93   &   19.45    &   47.13  &   41.74   &   37.19  &   -0.0572    &   0.0018   \\
        & ImageNet features &  50.24 & 21.48 &  32.25 &   50.49  &     33.09    &   37.80  &   -0.1446    &   0.0188  \\
        & CLIP ViT-B/16 &  51.66 & 27.97 &  26.24 &   44.46  &     35.85    &   35.70  &   0.0402    &   0.0168  \\
        & CLIP ViT-L/14 &  54.24 & 41.56 &  38.31 &   38.14  &     38.96    &   34.99  &   0.1672    &   0.0483  \\
        & AE2~\cite{ae2} &  {66.23} & {57.41} &  \underline{71.72} &   {65.85}  &     {64.59}    &   {62.15}  &   {0.5109}   &   {0.6316}   \\
        \cmidrule(lr){2-10}
        & \method (ResNet-50)  &  \underline{72.57} & {67.91} &  {70.74} &   \underline{68.42}  &     {63.27}    &   {63.85}  &   {0.7751}   &   {0.7463}   \\
        & \method (ViT-B/16)  &  \textbf{74.30} & \textbf{75.01} &  {71.28} &   {67.17}  &     \textbf{70.65}    &   \textbf{69.02}  &   \textbf{0.8533}   &   \textbf{0.9451}   \\
        & \method (ViT-L/14)  &  {72.41} & \underline{70.11} &  \textbf{72.92} &   \textbf{75.59}  &     \underline{67.73}    &   \underline{67.55}  &   \underline{0.8272}   &   \underline{0.8940}   \\
        \midrule
        \multirow{8}[2]{*}{(B)} & Random features &  36.84  &   33.96   &   41.97    &   52.48  &   50.56   &   51.98  &   -0.0477    &   0.0050   \\
        & ImageNet features &  41.59 & 39.93 &  45.52 &   54.09  &     27.31    &   43.21  &   -2.6681    &   0.0115  \\
        & CLIP ViT-B/16 &  43.24 & 49.21 &  30.94 &   52.16  &     46.39    &   40.34  &   -4.0754    &   0.0046  \\
        & CLIP ViT-L/14 &  46.65 & 46.79 &  17.77 &   46.20  &     44.32    &   53.75  &   -0.4735    &   0.0503  \\
        & AE2~\cite{ae2} &  85.17 & 84.73 &  82.77 &   84.90  &     78.48    &   \underline{83.41}  &   0.7634   &   0.9062   \\
        % \textsc{DeCap}~\citep{decap} &   &   &   \checkmark    & CLIP & CLIP & 50.6 &  -    \\
        \cmidrule(lr){2-10}
        % \textbf{Ours}   &    \checkmark     &   &    &  CLIP &    CLIP   &  \bf 58.3  &  \bf 81.7    \\
        % \textbf{Ours}   &    \checkmark     &   &    &  CLIP &    GPT-2   &   56.8  &   79.1    \\
        % \textbf{Ours}   &    \checkmark     &   &    &  BEiT &    GPT-2   &   53.2  &   71.7    \\
        % \textbf{Ours}   &    & \checkmark  &   \checkmark    &  CLIP &    CLIP   &  37.1  &  69.5    \\
        % \textbf{Ours}   &    & \checkmark  &   \checkmark    &  BEiT &    GPT-2   &  -  &  -    \\
        & \method (ResNet-50)  &  \textbf{86.84} & {83.83} &  \textbf{87.00} &   {87.17}  &     {79.27}    &   {79.87}  &   {0.8082}   &   {0.9152}   \\
        & \method (ViT-B/16)  &  {86.46} & \underline{85.09} &  \underline{86.61} &   \textbf{89.42}  &     \textbf{87.73}    &   \textbf{85.06}  &   \textbf{0.8992}   &   \textbf{0.9466}   \\
        & \method (ViT-L/14)  &  \underline{86.76} & \textbf{85.54} &  {86.58} &   \underline{87.35}  &     \underline{82.51}    &   {82.61}  &   \underline{0.8407}   &   \underline{0.9448}   \\
        \midrule
        \multirow{8}[2]{*}{(C)} & Random features &  45.26  &   47.45   &   44.33    &   49.83  &   55.44   &   55.75  &   -0.1303    &   -0.0072   \\
        & ImageNet features &  53.13 & 22.44 &  44.61 &   51.49  &     52.17    &   30.44  &   -1.6329    &   -0.0053  \\
        & CLIP ViT-B/16 &  60.60 & 36.97 &  48.43 &   43.63  &     47.58    &   37.02  &   -0.3139    &   -0.0048  \\
        & CLIP ViT-L/14 &  54.38 & 6.83 &  51.69 &  50.01   &   31.82      &  54.61   &   -0.2066    &   -0.0052  \\
        & AE2~\cite{ae2} &  66.56 & 57.15 &  65.60 &   {65.54}  &     65.79    &   57.35  &   0.1380   &   0.0934   \\
        % \textsc{DeCap}~\citep{decap} &   &   &   \checkmark    & CLIP & CLIP & 50.6 &  -    \\
        \cmidrule(lr){2-10}
        % \textbf{Ours}   &    \checkmark     &   &    &  CLIP &    CLIP   &  \bf 58.3  &  \bf 81.7    \\
        % \textbf{Ours}   &    \checkmark     &   &    &  CLIP &    GPT-2   &   56.8  &   79.1    \\
        % \textbf{Ours}   &    \checkmark     &   &    &  BEiT &    GPT-2   &   53.2  &   71.7    \\
        % \textbf{Ours}   &    & \checkmark  &   \checkmark    &  CLIP &    CLIP   &  37.1  &  69.5    \\
        % \textbf{Ours}   &    & \checkmark  &   \checkmark    &  BEiT &    GPT-2   &  -  &  -    \\
        & \method (ResNet-50)  &  {78.63} & \textbf{73.67} &  \underline{76.53} &   \textbf{71.47}  &     {66.74}    &   \underline{71.17}  &   {0.3982}   &   {0.2883}   \\
        & \method (ViT-B/16) &  \textbf{79.48} & \underline{71.83} &  {76.23} &   \underline{71.06}  &     \underline{75.03}    &   {70.03}  &   \underline{0.4483}   &   \underline{0.3052}   \\
        & \method (ViT-L/14)  &  \textbf{79.48} & {71.49} &  \textbf{76.61} &   {70.36}  &     \textbf{76.48}    &   \textbf{73.38}  &   \textbf{0.4534}   &   \textbf{0.3084}   \\
        \midrule
        \multirow{8}[2]{*}{(D)} & Random Features &  30.31  &   33.42   &   28.10    &   66.47  &   58.98   &   59.87  &   -0.0425    &   0.0177   \\
        & ImageNet Features &  69.15 & 42.03 &  58.61 &   76.96  &     66.90    &   60.31  &   -0.4143    &   0.0734  \\
        & CLIP ViT-B/16 &  67.81 & 43.41 &  44.22 &   74.54  &     59.57    &   52.02  &   -0.4996    &   0.0618  \\
        & CLIP ViT-L/14 &  64.40 & 47.53 &  47.50 &   74.26  &     67.19    &   58.73  &   -0.4126    &   0.0302  \\
        & AE2~\cite{ae2} &  85.87 & 84.71 & 85.68 & 86.83 & 81.46 & 82.07 & 0.5060 & 0.6171   \\
        % \textsc{DeCap}~\citep{decap} &   &   &   \checkmark    & CLIP & CLIP & 50.6 &  -    \\
        \cmidrule(lr){2-10}
        % \textbf{Ours}   &    \checkmark     &   &    &  CLIP &    CLIP   &  \bf 58.3  &  \bf 81.7    \\
        % \textbf{Ours}   &    \checkmark     &   &    &  CLIP &    GPT-2   &   56.8  &   79.1    \\
        % \textbf{Ours}   &    \checkmark     &   &    &  BEiT &    GPT-2   &   53.2  &   71.7    \\
        % \textbf{Ours}   &    & \checkmark  &   \checkmark    &  CLIP &    CLIP   &  37.1  &  69.5    \\
        % \textbf{Ours}   &    & \checkmark  &   \checkmark    &  BEiT &    GPT-2   &  -  &  -    \\
        & \method (ResNet-50)  &  \underline{89.34} & \textbf{94.83} &  {84.96} &   {89.83}  &     {86.71}    &   {82.68}  &   {0.7588}   &   {0.7599}   \\
        & \method (ViT-B/16)  &  {89.12} & {94.47} &  \underline{85.73} &   \underline{90.61}  &     \textbf{88.34}    &   \textbf{88.94}  &   \textbf{0.7881}   &   \underline{0.7852}   \\
        & \method (ViT-L/14)  &  \textbf{89.56} & \underline{94.48} &  \textbf{86.51} &   \textbf{91.21}  &     \underline{87.04}    &   \underline{88.33}  &   \underline{0.7653}   &   \textbf{0.8101}   \\
        \bottomrule
        \end{tabular}
        % \vspace{-5pt}
    \end{table*}
\begin{table*}[t]
    \centering
    \small
    \caption{Performance comparison for few-shot classification and regular frame retrieval on the AE2 benchmark~\cite{ae2}. The benchmark consists of four sub-tasks: (A) Break Eggs, (B) Pour Milk, (C) Pour Liquid, and (D) Tennis Forehand. We report the few-shot classification (F1 score) and regular frame retrieval (mAP@5, mAP@10, and mAP@15) performance. The top results are highlighted in \textbf{bold} and the second-best results are \underline{underlined}.
    }\label{tab:fewshot}
    \setlength{\tabcolsep}{4pt}
        \begin{tabular}{c l >{\centering\arraybackslash}p{1.2cm} >{\centering\arraybackslash}p{1.2cm} >{\centering\arraybackslash}p{1.2cm} ccc}
        \toprule
        \multirow{2}[2]{*}{Task} & \multirow{2}[2]{*}{Method} & \multicolumn{3}{c}{Few-shot Classification (F1 score)} & \multicolumn{3}{c}{Regular Frame Retrieval}  \\ 
        \cmidrule(lr){3-5} \cmidrule(lr){6-8}
        &           &       10\%    &   50\%    &   100\%   &   mAP@5   &   mAP@10   &   mAP@15                \\       
        \midrule
        \multirow{13}[2]{*}{(A)} & Random features &  19.18  &   19.18   &   19.18   &   48.26  &   47.13   &   45.75   \\
        & ImageNet features & 46.15  & 48.80 &  50.24 &  49.98   &     50.49    &  50.08    \\
        & CLIP ViT-B/16 & 46.46  & 49.18 &  51.66 & 44.89    &     44.46    &   43.44   \\
        & CLIP ViT-L/14 & 47.36  & 51.80 &  54.24 &  38.47   &     38.14    &   37.72   \\
        & ActorObserverNet~\cite{aon} & 31.40  & 35.63 &  36.14 & 50.92 & 50.47 & 49.72     \\
        & TCN~\cite{tcn} (single-view) & 52.30  & 54.90 &  56.90 &  52.82   & 53.42 & 53.60     \\
        & TCN~\cite{tcn} (multi-view) & 56.88  & 59.25 & 59.91 & 59.11 & 58.83 & 58.44     \\
        & TCN~\cite{tcn} (unpaired multi-view) & 56.13 & 56.65 & 56.79 & 58.18 & 57.78 & 57.21     \\
        & CARL~\cite{carl} & 39.18 & 41.92 & 43.43 & 47.14 & 46.04 & 44.99     \\
        & TCC~\cite{tcc} & 57.54 & 59.18 & 59.84 & 59.33 & 58.75 & 57.99     \\
        & GTA~\cite{gta} & 56.89 & 56.77 & 56.86 & 62.79 & 61.55 & 60.38     \\
        & AE2~\cite{ae2} & \underline{63.95} & \underline{64.86} &  \underline{66.23} & \underline{66.86} & \underline{65.85} & \underline{64.73}   \\
        \cmidrule(lr){2-8}
        & \method (ViT-B/16)  &  \textbf{71.84} & \textbf{73.92} &  \textbf{74.30} &   \textbf{67.28}  &     \textbf{67.17}    &   \textbf{66.40}    \\
        \midrule
        \multirow{10}[2]{*}{(B)} & Random features &  36.84  &   33.96   &   41.97    &   52.48  &   50.56   &   51.98   \\
        & ImageNet features &  41.59 & 39.93 &  45.52 &   54.09  &     27.31    &   43.21    \\
        & CLIP ViT-B/16 & 39.44  & 38.90 & 43.24 &  53.29   &     52.16    &  51.55    \\
        & CLIP ViT-L/14 & 42.68 & 39.91 & 46.65 & 46.20   &      46.20    & 53.75     \\
        & TCN~\cite{tcn} (single-view) & 43.60 & 46.83 & 47.39 & 56.98 & 57.00 & 56.46     \\
        & CARL~\cite{carl} & 48.73 & 48.78 & 48.79 & 55.29 & 55.01 & 54.23     \\
        & TCC~\cite{tcc} & 78.69 & 77.97 & 77.91 & 81.22 & 80.97 & 80.46     \\
        & GTA~\cite{gta} & 79.82 & 80.96 & 81.11 & 80.65 & 80.12 & 79.68     \\
        & AE2~\cite{ae2} & \underline{85.17} & \underline{85.12} & \underline{85.17} & \underline{85.25} & \underline{84.90} & \underline{84.55}   \\
        % \textsc{DeCap}~\citep{decap} &   &   &   \checkmark    & CLIP & CLIP & 50.6 &  -    \\
        \cmidrule(lr){2-8}
        % \textbf{Ours}   &    \checkmark     &   &    &  CLIP &    CLIP   &  \bf 58.3  &  \bf 81.7    \\
        % \textbf{Ours}   &    \checkmark     &   &    &  CLIP &    GPT-2   &   56.8  &   79.1    \\
        % \textbf{Ours}   &    \checkmark     &   &    &  BEiT &    GPT-2   &   53.2  &   71.7    \\
        % \textbf{Ours}   &    & \checkmark  &   \checkmark    &  CLIP &    CLIP   &  37.1  &  69.5    \\
        % \textbf{Ours}   &    & \checkmark  &   \checkmark    &  BEiT &    GPT-2   &  -  &  -    \\
        & \method (ViT-B/16)  &  \textbf{86.12} & \textbf{86.44} &  \textbf{86.46} &   \textbf{90.99}  &     \textbf{89.42}    &   \textbf{88.98}   \\
        \midrule
        \multirow{10}[2]{*}{(C)} & Random features &  45.26  &   47.45   &   44.33    &   49.83  &   55.44   &   55.75    \\
        & ImageNet features &  53.13 & 22.44 &  44.61 &   51.49  &     52.17    &   30.44   \\
        & CLIP ViT-B/16 & 57.21  & 35.46 &  60.60 & 42.34 & 43.63 & 44.03   \\
        & CLIP ViT-L/14 & 51.72  & 28.30 &  54.38 &  48.56   &    50.01     &  50.52    \\
        & TCN~\cite{tcn} (single-view) & 54.62 & 55.08 & 54.02 & 48.50 & 48.83 & 49.03     \\
        & CARL~\cite{carl} & 51.68 & 55.67 & 56.98 & 55.03 & 55.29 & 54.93     \\
        & TCC~\cite{tcc} & 52.37 & 51.70 & 52.53 & 62.93 & 62.33 & 61.44     \\
        & GTA~\cite{gta} & 55.91 & 56.87 & 56.92 & 62.83 & 62.79 & 62.12     \\
        & AE2~\cite{ae2} & \underline{65.88} & \underline{66.53} & \underline{66.56} & \underline{66.55} & \underline{65.54} & \underline{64.66}   \\
        % \textsc{DeCap}~\citep{decap} &   &   &   \checkmark    & CLIP & CLIP & 50.6 &  -    \\
        \cmidrule(lr){2-8}
        % \textbf{Ours}   &    \checkmark     &   &    &  CLIP &    CLIP   &  \bf 58.3  &  \bf 81.7    \\
        % \textbf{Ours}   &    \checkmark     &   &    &  CLIP &    GPT-2   &   56.8  &   79.1    \\
        % \textbf{Ours}   &    \checkmark     &   &    &  BEiT &    GPT-2   &   53.2  &   71.7    \\
        % \textbf{Ours}   &    & \checkmark  &   \checkmark    &  CLIP &    CLIP   &  37.1  &  69.5    \\
        % \textbf{Ours}   &    & \checkmark  &   \checkmark    &  BEiT &    GPT-2   &  -  &  -    \\
        & \method (ViT-B/16) &  \textbf{79.10} & \textbf{79.28} &  \textbf{79.48} &   \textbf{73.89}  &     \textbf{71.06}    &   \textbf{67.83} \\
        \midrule
        \multirow{10}[2]{*}{(D)} & Random Features &  30.31  &   33.42   &   28.10    &   66.47  &   58.98   &   59.87   \\
        & ImageNet Features &  69.15 & 42.03 &  58.61 &   76.96  &     66.90    &   60.31   \\
        & CLIP ViT-B/16 & 70.37  & 48.01 &  67.81 &  76.45   &     74.54    &  73.15    \\
        & CLIP ViT-L/14 & 68.44 & 42.95 &  64.40 &  75.61   &     74.26    &   73.14   \\
        & TCN~\cite{tcn} (single-view) & 65.78  & 69.19 & 68.87 & 74.05 & 73.76 & 73.10     \\
        & CARL~\cite{carl} & 58.89 & 59.38 & 59.69 & 72.94 & 69.43 & 67.14     \\
        & TCC~\cite{tcc} & 67.71 & 77.07 & 78.41 & 82.78 & 80.24 & 78.59     \\
        & GTA~\cite{gta} & 80.31 & 83.04 & 83.63 & 86.59 & 85.20 & 84.33     \\
        & AE2~\cite{ae2} & \underline{85.24} & \underline{85.72} & \underline{85.87} & \underline{87.94} & \underline{86.83} & \underline{86.05}  \\
        % \textsc{DeCap}~\citep{decap} &   &   &   \checkmark    & CLIP & CLIP & 50.6 &  -    \\
        \cmidrule(lr){2-8}
        % \textbf{Ours}   &    \checkmark     &   &    &  CLIP &    CLIP   &  \bf 58.3  &  \bf 81.7    \\
        % \textbf{Ours}   &    \checkmark     &   &    &  CLIP &    GPT-2   &   56.8  &   79.1    \\
        % \textbf{Ours}   &    \checkmark     &   &    &  BEiT &    GPT-2   &   53.2  &   71.7    \\
        % \textbf{Ours}   &    & \checkmark  &   \checkmark    &  CLIP &    CLIP   &  37.1  &  69.5    \\
        % \textbf{Ours}   &    & \checkmark  &   \checkmark    &  BEiT &    GPT-2   &  -  &  -    \\
        & \method (ViT-B/16)  &  \textbf{88.78} & \textbf{89.01} &  \textbf{89.12} &   \textbf{90.89}  &     \textbf{90.61}    &   \textbf{90.87}   \\
        \bottomrule
        \end{tabular}
        \vspace{5pt}
    \end{table*}

\begin{table*}[t]
    \centering
    \small
    \caption{Performance comparison for cross-view retrieval on the AE2 benchmark~\cite{ae2}. The benchmark consists of four sub-tasks: (A) Break Eggs, (B) Pour Milk, (C) Pour Liquid, and (D) Tennis Forehand. We report the cross-view frame retrieval (mAP@5, mAP@10, and mAP@15) performance. The top results are highlighted in \textbf{bold} and the second-best results are \underline{underlined}.
    }\label{tab:retrieval}
    \setlength{\tabcolsep}{4pt}
        \begin{tabular}{clcccccc}
        \toprule
        \multirow{2}[2]{*}{Task} & \multirow{2}[2]{*}{Method} & \multicolumn{3}{c}{Ego2Exo Frame Retrieval} & \multicolumn{3}{c}{Exo2Ego Frame Retrieval}  \\ 
        \cmidrule(lr){3-5} \cmidrule(lr){6-8}
       &  & mAP@5    & mAP@10   &   mAP@15    & mAP@5    & mAP@10   &   mAP@15                \\       
        \midrule
        \multirow{10}[2]{*}{(A)} & Random features & 42.51 & 41.74 & 40.51 & 38.08 & 38.19 & 37.10   \\
        & ImageNet features & 33.32 & 33.09 & 32.78 & 38.99 & 37.80 & 36.71    \\
        & CLIP ViT-B/16 & 35.80 & 35.85 & 34.92 & 34.91 & 35.70 &  35.96     \\
        & CLIP ViT-L/14 & 39.30 & 38.94 & 38.14 & 35.23 & 34.99 & 33.98     \\
        & ActorObserverNet~\cite{aon} & 43.57 & 42.70 & 41.56 & 42.00 & 41.29 & 40.48     \\
        & TCN~\cite{tcn} (single-view) & 31.12 & 32.63 & 33.73 & 34.67 & 34.91 & 35.31     \\
        & TCN~\cite{tcn} (multi-view) & 46.38 & 47.04 & 46.96 & 52.50 & 52.68 & 52.43     \\
        & TCN~\cite{tcn} (unpaired multi-view) & 55.34 & 54.64 & 53.75 & 58.79 & 57.87 & 57.07     \\
        & CARL~\cite{carl} & 37.89 & 37.38 & 36.57 & 40.37 & 39.94 & 39.38     \\
        & TCC~\cite{tcc} & 62.11 & 61.11 & 60.33 & 62.39 & 62.03 & 61.25     \\
        & GTA~\cite{gta} & 57.11 & 56.25 & 55.10 & 54.47 & 53.93 & 53.22     \\
        & AE2~\cite{ae2} & \underline{65.70} & \underline{64.59} & \underline{63.76} & \underline{62.48} & \underline{62.15} & \underline{61.80}   \\
        \cmidrule(lr){2-8}
        & \method (ViT-B/16)  &  \textbf{72.76} & \textbf{70.65} &  \textbf{70.27} &   \textbf{71.79}  &     \textbf{69.02}    &   \textbf{68.94}    \\
        \midrule
        \multirow{8}[2]{*}{(B)} & Random features &  51.46 & 50.56 & 48.93  & 52.78 & 51.98 & 50.82   \\
        & ImageNet features & 25.72 & 27.31 & 28.57 & 41.50 & 43.21 & 43.06    \\
        & CLIP ViT-B/16 & 46.37 & 46.39 & 46.86 & 41.28 & 40.34 &  39.86    \\
        & CLIP ViT-L/14 & 43.71 & 44.32 & 44.20 & 55.55 & 53.75 & 53.10  \\
        & TCN~\cite{tcn} (single-view) & 47.00 & 46.48 & 45.42 & 47.94 & 47.20 & 46.59     \\
        & CARL~\cite{carl} & 54.35 & 52.99 & 51.99 & 51.14 & 51.51 & 51.00     \\
        & TCC~\cite{tcc} & 75.54 & 75.30 & 75.02 & 80.44 & 80.27 & 80.18     \\
        & GTA~\cite{gta} & 72.55 & 72.78 & 72.96 & 75.16 & 75.40 & 75.48     \\
        & AE2~\cite{ae2} & \underline{78.21} & \underline{78.48} & \underline{78.78} & \underline{83.88} & \underline{83.41} & \underline{83.05}   \\
        % \textsc{DeCap}~\citep{decap} &   &   &   \checkmark    & CLIP & CLIP & 50.6 &  -    \\
        \cmidrule(lr){2-8}
        % \textbf{Ours}   &    \checkmark     &   &    &  CLIP &    CLIP   &  \bf 58.3  &  \bf 81.7    \\
        % \textbf{Ours}   &    \checkmark     &   &    &  CLIP &    GPT-2   &   56.8  &   79.1    \\
        % \textbf{Ours}   &    \checkmark     &   &    &  BEiT &    GPT-2   &   53.2  &   71.7    \\
        % \textbf{Ours}   &    & \checkmark  &   \checkmark    &  CLIP &    CLIP   &  37.1  &  69.5    \\
        % \textbf{Ours}   &    & \checkmark  &   \checkmark    &  BEiT &    GPT-2   &  -  &  -    \\
        & \method (ViT-B/16)  &  \textbf{85.15} & \textbf{87.73} &  \textbf{87.80} &   \textbf{85.48}  &     \textbf{85.06}    &   \textbf{85.00}   \\
        \midrule
        \multirow{8}[2]{*}{(C)} & Random features &  55.78 &   55.44   &   54.77    &   56.31  &   55.75   &   54.56    \\
        & ImageNet features &  51.44 & 52.17 &  52.38 & 30.18 &  30.44   &   30.40   \\
        & CLIP ViT-B/16 & 42.08 & 47.58 & 49.78 & 35.14 & 37.02 & 36.71  \\
        & CLIP ViT-L/14 & 32.33 & 31.82 & 31.59 & 54.01 & 54.61 & 54.64  \\
        & TCN~\cite{tcn} (single-view) & 53.60 & 55.28 & 55.46 & 29.16 & 31.15 & 31.95     \\
        & CARL~\cite{carl} & 59.59 & 59.37 & 59.19 & 34.73 & 36.80 & 38.10     \\
        & TCC~\cite{tcc} & 55.98 & 56.08 & 56.13 & \underline{58.11} & \underline{57.89} & \underline{57.15}     \\
        & GTA~\cite{gta} & 57.03 & 58.52 & 59.00 & 51.71 & 53.32 & 53.54     \\
        & AE2~\cite{ae2} & \underline{66.23} & \underline{65.79} & \underline{65.00} & 57.42 & 57.35 & 57.03   \\
        % \textsc{DeCap}~\citep{decap} &   &   &   \checkmark    & CLIP & CLIP & 50.6 &  -    \\
        \cmidrule(lr){2-8}
        % \textbf{Ours}   &    \checkmark     &   &    &  CLIP &    CLIP   &  \bf 58.3  &  \bf 81.7    \\
        % \textbf{Ours}   &    \checkmark     &   &    &  CLIP &    GPT-2   &   56.8  &   79.1    \\
        % \textbf{Ours}   &    \checkmark     &   &    &  BEiT &    GPT-2   &   53.2  &   71.7    \\
        % \textbf{Ours}   &    & \checkmark  &   \checkmark    &  CLIP &    CLIP   &  37.1  &  69.5    \\
        % \textbf{Ours}   &    & \checkmark  &   \checkmark    &  BEiT &    GPT-2   &  -  &  -    \\
        & \method (ViT-B/16) &  \textbf{79.06} & \textbf{75.03} &  \textbf{72.73} &   \textbf{76.21}  &     \textbf{70.03}    &   \textbf{69.44} \\
        \midrule
        \multirow{8}[2]{*}{(D)} & Random Features &  61.24  &   58.98   &   56.94    &   63.42  &   59.87   &   57.57   \\
        & ImageNet Features &  69.34 & 66.90 & 64.95 & 61.61 & 60.31 & 58.55   \\
        & CLIP ViT-B/16 & 60.63 & 59.57 & 58.46 & 52.25 & 52.02 & 52.12 \\
        & CLIP ViT-L/14 & 69.02 & 67.19 & 65.44 & 61.83 & 58.73 & 57.05 \\
        & TCN~\cite{tcn} (single-view) & 54.12 & 55.08 & 55.05 & 56.70 & 56.65 & 55.84     \\
        & CARL~\cite{carl} & 52.18 & 54.83 & 55.39 & 65.94 & 63.19 & 60.83     \\
        & TCC~\cite{tcc} & 57.87 & 55.84 & 53.81 & 48.62 & 47.27 & 46.11     \\
        & GTA~\cite{gta} & 78.93 & 78.00 & 77.01 & 79.95 & 79.14 & 78.52     \\
        & AE2~\cite{ae2} & \underline{82.58} & \underline{81.46} & \underline{80.75} & \underline{82.82} & \underline{82.07} & \underline{81.69}  \\
        % \textsc{DeCap}~\citep{decap} &   &   &   \checkmark    & CLIP & CLIP & 50.6 &  -    \\
        \cmidrule(lr){2-8}
        % \textbf{Ours}   &    \checkmark     &   &    &  CLIP &    CLIP   &  \bf 58.3  &  \bf 81.7    \\
        % \textbf{Ours}   &    \checkmark     &   &    &  CLIP &    GPT-2   &   56.8  &   79.1    \\
        % \textbf{Ours}   &    \checkmark     &   &    &  BEiT &    GPT-2   &   53.2  &   71.7    \\
        % \textbf{Ours}   &    & \checkmark  &   \checkmark    &  CLIP &    CLIP   &  37.1  &  69.5    \\
        % \textbf{Ours}   &    & \checkmark  &   \checkmark    &  BEiT &    GPT-2   &  -  &  -    \\
        & \method (ViT-B/16)  &  \textbf{88.55} & \textbf{88.34} &  \textbf{87.98} &   \textbf{90.64}  &     \textbf{88.94}    &   \textbf{87.26}   \\
        \bottomrule
        \end{tabular}
        \vspace{20pt}
    \end{table*}
    
\begin{figure*}[t]
  \centering
    \begin{subfigure}{0.48\linewidth}
    \centering
    {\includegraphics[width=1.0\linewidth]{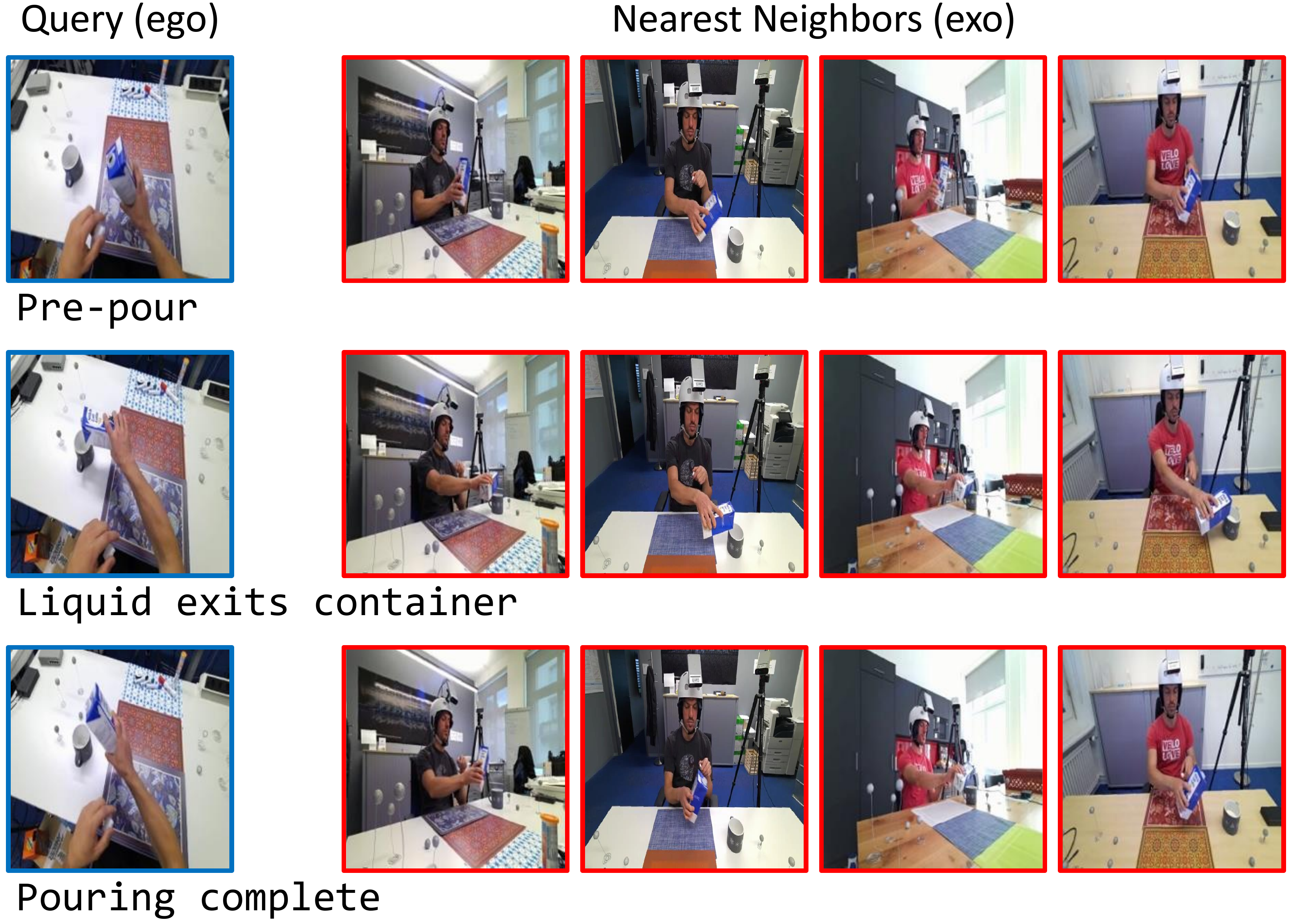}}
    \caption{Ego2Exo frame retrieval on Pour Milk}\label{fig:retrievala}
   \end{subfigure}  \hfill % \\ \vspace{5pt}
   \begin{subfigure}{0.48\linewidth}
    \centering
    {\includegraphics[width=1.0\linewidth]{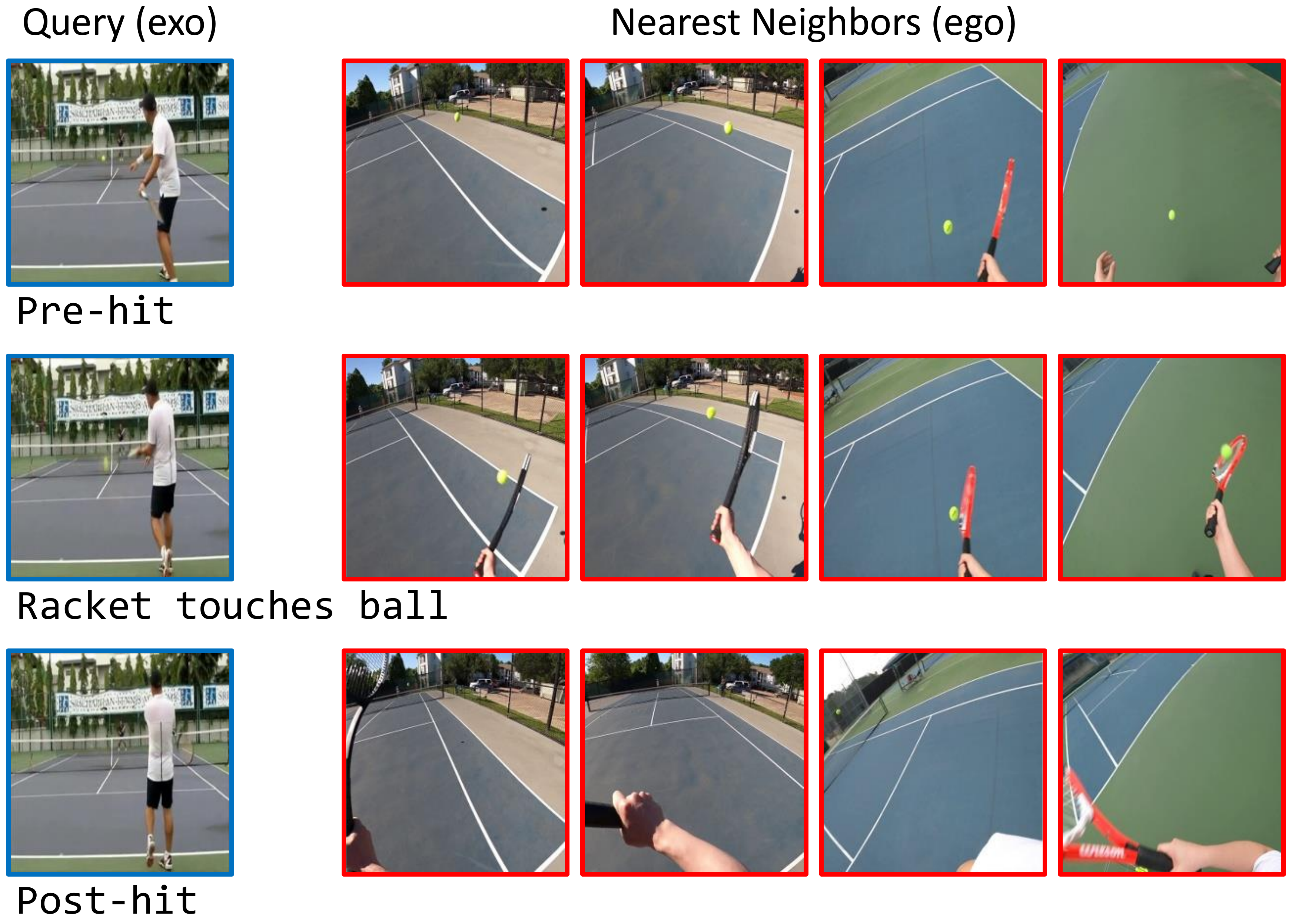}}
    \caption{Exo2Ego frame retrieval on Tennis Forehand}\label{fig:retrievalb}
   \end{subfigure}
  \caption{Qualitative examples of frame retrieval from the Pour Milk and Tennis Forehand datasets. We retrieve the nearest neighbor frames (red box) corresponding to the given query frame (blue box).}
  \label{fig:retrieval}
\end{figure*}
    
\begin{table*}[t]
    \centering
    \small
    \caption{Performance comparison according to various sizes of latent space in \method. We evaluate the performance on the Break Eggs dataset.
    }\label{tab:latent}
    \setlength{\tabcolsep}{4pt}
        \begin{tabular}{cccccccccc}
        \toprule
        \multirow{2}[2]{*}{\makecell{Latent \\ Size}} & \multirow{2}[2]{*}{\makecell{Trainable \\ Params}} &\multicolumn{3}{c}{Classification (F1 score)} & \multicolumn{3}{c}{Frame Retrieval (mAP@10)}& \multirow{2}[2]{*}{\makecell{ Phase\\ progression} } & \multirow{2}[2]{*}{\makecell{ Kendall's \\ $\tau$} } \\ 
         \cmidrule(lr){3-5} \cmidrule(lr){6-8}
        &  & Regular  & Ego2Exo & Exo2Ego    & Regular    & Ego2Exo   &   Exo2Ego   &       &                \\       
        \midrule
        64  & 0.9M & \underline{71.55} & 71.34 & 69.36 & 65.87 & 64.09 & 68.16 & 0.8362 & 0.8943 \\
        128 & 3.4M & 70.84 & \underline{72.74} & \underline{69.71} & 67.07 & 66.70 & 68.45 & \underline{0.8407} & \underline{0.9240} \\
        256 & 12.3M &  \textbf{74.30} & \textbf{75.01} &  \textbf{71.28} &   \underline{67.17}  &     \underline{70.65}    &   \underline{69.02}  &   \textbf{0.8533}   &   \textbf{0.9451}   \\
        512 & 51.5M & 70.89 & 70.19 & 68.72 & \textbf{68.70} & \textbf{73.29} & \textbf{74.45} & 0.8107 & \underline{0.9240} \\ 
        \bottomrule
        \end{tabular}
    \end{table*}
    % In this section, we provide additional experimental results, including the performance analysis of \method with the CLIP pre-trained ViT-L/14, ablation studies corresponding to each hyper-parameter, and qualitative results.
\subsection{Results with different frame encoders}
    We mainly used the CLIP pretrained ViT-B/16~\cite{vit} to encode each frame in the main paper.
    To demonstrate the robustness of \method according to the frame encoders, we train \method with the CLIP pretrained ViT-L/14~\cite{vit} and ResNet-50~\cite{resnet}, and evaluate the performance on four tasks for four datasets.
    Implementation details for each frame encoder are as follows.
    \begin{itemize}
        \item {\textbf{CLIP ViT-L/14}~\cite{vit} pretrained on LAION-400M~\cite{laion-400m} projects each frame into 1024-dimensional 256 token embeddings different from the ViT-B/16, which has 768-dimensional 196 token embeddings.
        % Therefore, the decoder $h_\psi(\cdot)$ may struggle to reconstruct the original token embeddings from the lower dimensional latent space (\ie, 256-dim as in \method with the ViT-B/16).
        We keep the number of layers of the encoder $g_\phi(\cdot)$ and the decoder $h_\psi(\cdot)$ as 12 and 4 while setting the size of the latent space to 512.
        The number of trainable parameters is 51.8M (38.4M for the encoder and 13.4M for the decoder, respectively).
        The token selection ratio in selective token merging (STM), and the masking ratio in masked self-view modeling (MSM) and masked cross-view modeling (MCM) are set to 0.3, 0.4, and 0.8 as with the ViT-B/16.
        }
        \item {\textbf{ResNet-50}~\cite{resnet} employs convolutional neural network, and is pretrained on ImageNet-1K~\cite{imagenet}.
        We extract the feature for each frame from a \textit{Conv4c} layer of ResNet-50, which has $14\times14$ resolution with 1024 dimensions.
        We also perform the selective token merging to keep the overall framework of \method.
        The receptive field of each 1024-d embedding is $55\times55$ pixels, which is wider than $16\times16$ in ViT-B/16.
        Therefore, we reduce the selection ratio to 0.1.
        The masking ratio in MSM and MCM are set to 0.4 and 0.8, respectively.
        Similar to the ViT-L/14, we use 512-dimensional latent space for the encoder $g_\phi(\cdot)$ and the decoder $h_\psi(\cdot)$.
        }
    \end{itemize}
    In \cref{tab:vitL}, we first provide the zero-shot performance of the ResNet-50 (ImageNet features), CLIP ViT-B/16, and CLIP ViT-L/14.
    While ViT-L/14 (303M) has about three times more parameters than ViT-B/16 (86M), comparisons between the two frame encoders show that generalization capability is not dependent on the model size.
    Meanwhile, our \method with various frame encoders consistently outperforms the state-of-the-art~\cite{ae2} across tasks and datasets.
    In practice, \method with the ResNet-50 surpasses AE2~\cite{ae2} without any additional information such as bounding boxes from the hand-object detector as in \cite{ae2}.
    It demonstrates the robustness of the framework of our \method.

    \begin{figure*}[t]
      \centering
        \begin{subfigure}{1.0\linewidth}
        \centering
        {\includegraphics[width=1.0\linewidth]{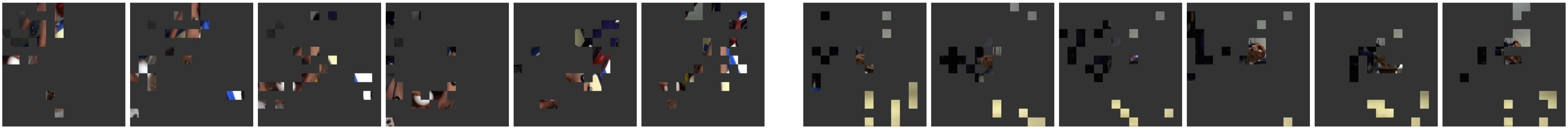}}
        \caption{Selected tokens with ratio 0.1 in ego (left) and exo (right) videos.}\label{fig:stma}
       \end{subfigure}  \\ \vspace{3pt}
       \begin{subfigure}{1.0\linewidth}
        \centering
        {\includegraphics[width=1.0\linewidth]{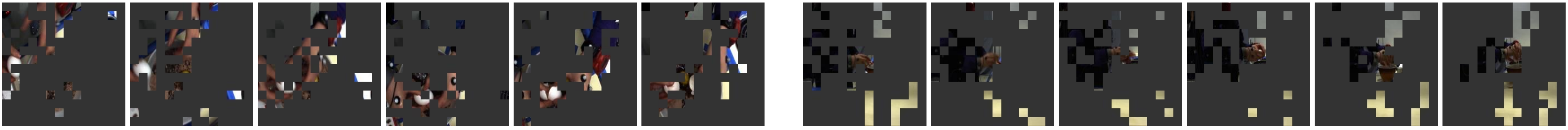}}
        \caption{Selected tokens with ratio 0.2 in ego (left) and exo (right) videos.}\label{fig:stmb}
       \end{subfigure}  \\ \vspace{3pt}
       \begin{subfigure}{1.0\linewidth}
        \centering
        {\includegraphics[width=1.0\linewidth]{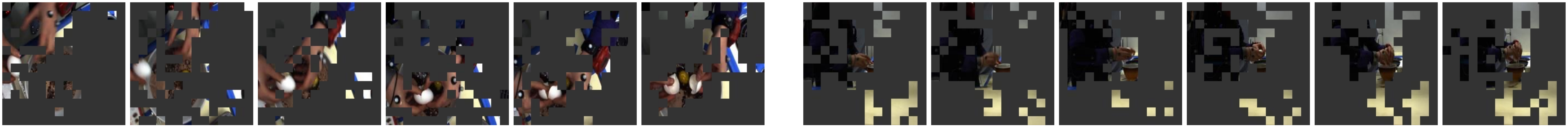}}
        \caption{Selected tokens with ratio 0.3 in ego (left) and exo (right) videos.}\label{fig:stmc}
       \end{subfigure}  \\ \vspace{3pt}
       \begin{subfigure}{1.0\linewidth}
        \centering
        {\includegraphics[width=1.0\linewidth]{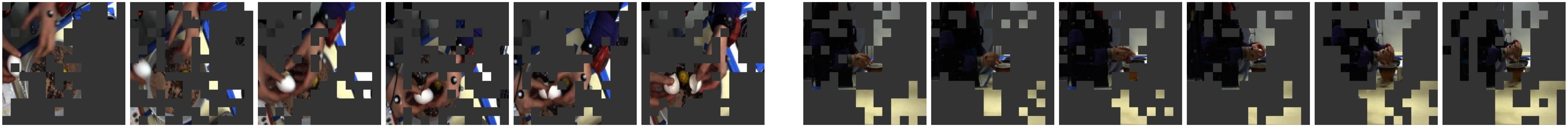}}
        \caption{Selected tokens with ratio 0.4 in ego (left) and exo (right) videos.}\label{fig:stmd}
       \end{subfigure}  \\ \vspace{3pt}
       \begin{subfigure}{1.0\linewidth}
        \centering
        {\includegraphics[width=1.0\linewidth]{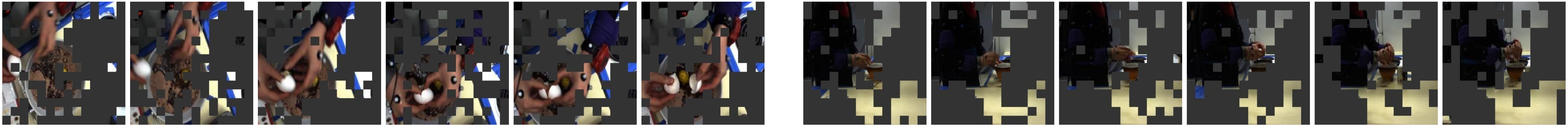}}
        \caption{Selected tokens with ratio 0.5 in ego (left) and exo (right) videos.}\label{fig:stme}
       \end{subfigure}  \\ \vspace{3pt}
       \begin{subfigure}{1.0\linewidth}
        \centering
        {\includegraphics[width=1.0\linewidth]{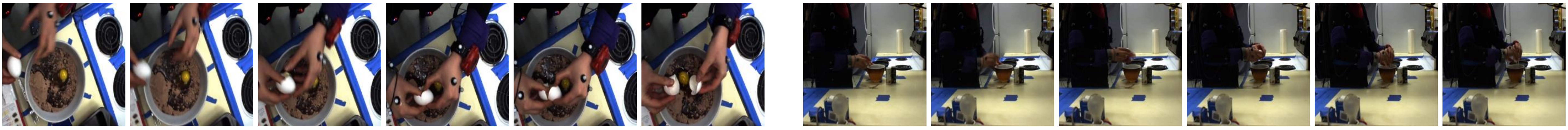}}
        \caption{Complete frames from ego (left) and exo (right) videos.}\label{fig:stmf}
       \end{subfigure}
      \caption{Visualization of selected tokens at each frame sampled from ego (left) and exo (right) videos. Note that complete frames are identical with the token selection ratio of 1.0.}
      \label{fig:stm}
    \end{figure*}
    
\subsection{Few-shot classification}
    Following \cite{ae2}, we compare few-shot classification performance with the state-of-the-art methods~\cite{aon, tcn, carl, tcc, gta, ae2} in \cref{tab:fewshot}.
    We train the SVM classifier using 10\% (or 50\%) of the latents from the training data and evaluate the classification performance.
    Note that we train \method ten times on non-overlapped few-shot training data and report the average performance.
    \cref{tab:fewshot} demonstrates the superior performance of \method, showing significant performance gaps to the existing works across all datasets.
    In particular, \method trained with only 10\% training data significantly outperforms the prior best performance~\cite{ae2} trained with 100\% training data by a large margin of 12.54 on the Pour Liquid dataset.
\subsection{Frame retrieval}\label{sec:retrieval_supp}
    In \cref{tab:fewshot} and \cref{tab:retrieval}, we report the frame retrieval performance, evaluated in both regular and cross-view settings.
    Comparisons with the existing methods consistently demonstrate the effectiveness of \method in both regular and cross-view retrieval across all datasets, showing an average performance improvement of about 10\%.
    
    We illustrate examples of cross-view frame retrieval from the Pour Milk and Tennis Forehand datasets in \cref{fig:retrieval}.
    Given the query frame (blue box) from one view, we retrieve the frames (red box) from the other view videos using NN search.
    The results show that the query and retrieved frames are contextually well-aligned through the action states.
    In addition, properly retrieved frames demonstrate that \method captures contexts over time.
    For example, the frames with the action phases of `pre-pour' and `pouring complete' are visually similar, however, \method successfully performs frame retrieval by capturing the context with respect to the action state over time.
    In this regard, we further analyze the effectiveness of \method by visualizing the frame embeddings in the following section.
\subsection{Ablation study}\label{sec:ablation_append}
    We analyze the effectiveness of each component in \method, including the size of latent space, token selection ratio in STM, and masking ratio in MSM and MCM.
    Note that we use the CLIP pretrained ViT-B/16 as the frame encoder for the following experiments.

\begin{table*}[t]
    \centering
    \small
    \caption{Performance comparison according to variants of the hyperparameters in \method. We report the performance evaluated on the Break Eggs dataset.
    }\label{tab:ratio}
    \setlength{\tabcolsep}{4pt}
        \begin{tabular}{ccccccccccc}
        \toprule
        \multicolumn{3}{c}{Ratio (\%)} &\multicolumn{3}{c}{Classification (F1 score)} & \multicolumn{3}{c}{Frame Retrieval (mAP@10)}& \multirow{2}[2]{*}{\makecell{ Phase\\ progression} } & \multirow{2}[2]{*}{\makecell{ Kendall's \\ $\tau$} } \\ 
        \cmidrule(lr){1-3} \cmidrule(lr){4-6} \cmidrule(lr){7-9}
        STM  & MSM & MCM & Regular  & Ego2Exo & Exo2Ego    & Regular    & Ego2Exo   &   Exo2Ego   &       &                \\       
        \midrule
        \multicolumn{5}{l}{\hspace{-5pt}\textit{Effectiveness of token selection ratio}} \\ 
        10  &   40    &   80  & 41.45 & 21.13 & 20.03 & 56.05 & 46.06 & 46.85 & 0.1858 & 0.0157   \\
        20  &   40    &   80  & 70.97 & 69.60 & 66.27 & 65.05 & 71.13 & 64.52 & 0.6597 & 0.7978   \\
        30  &   40    &   80  &  \textbf{74.30} & \textbf{75.01} &  \textbf{71.28} &   {67.17}  &     {70.65}    &   \textbf{69.02}  &   \textbf{0.8533}   &   \textbf{0.9451}   \\
        40  &   40    &   80  & \underline{72.39} & \underline{72.59} & \underline{69.19} & \textbf{68.20} & \textbf{73.79} & {67.44} & \underline{0.8299} & \underline{0.8963}   \\
        50  &   40    &   80  & 71.56 & 69.05 & 68.79 & \textbf{68.20} & \underline{72.79} & 67.20 & \underline{0.8299} & 0.8926  \\
        100  &   40    &   80  & 71.34 & 72.58 & 65.07 & \underline{67.44} & 69.32 & \underline{67.87} & 0.7894 & 0.8957   \\
        \midrule
        \multicolumn{5}{l}{\hspace{-5pt}\textit{Effectiveness of masking ratio in MSM}} \\ 
        30  &   10    &   80  & 70.71 & 69.51 & 66.20 & 67.67 & 66.10 & 63.89 & 0.5228 & 0.6724   \\
        30  &   20    &   80  & 71.22 & 70.38 & 69.81 & 67.67 & 68.27 & 65.83 & 0.8134 & 0.9126 \\
        30  &   30    &   80  & 72.28 & 73.21 & 70.22 & \textbf{67.28} & 70.21 & 68.15 & 0.8330 & 0.9337   \\
        30  &   40    &   80  &  \textbf{74.30} & \textbf{75.01} & \textbf{71.28} & \underline{67.17} & \underline{70.65} & \underline{69.02} & \textbf{0.8533} & \textbf{0.9451}   \\
        30  &   50    &   80  & \underline{72.87} & \underline{73.71} & \underline{70.87} & \textbf{67.28} & \textbf{71.21} & \textbf{70.15} &  \underline{0.8398} & \underline{0.9410}  \\
        30  &   100    &   80  & 66.65 & 69.97 & 68.24 & 65.01 & 67.48 & 66.86 & 0.6916 & 0.7818   \\
        \midrule        
        \multicolumn{5}{l}{\hspace{-5pt}\textit{Effectiveness of masking ratio in MCM}} \\ 
        30  &   40    &   0  & 67.23 & 66.65 & 67.10 & 60.38 & 58.44 & 56.97 & 0.7019 & 0.8040   \\
        30  &   40    &   20  & 71.40 & 69.06 & 70.19 & 64.98 & 62.09 & 61.27 & 0.8269 & 0.9112   \\
        30  &   40    &   40  & 73.23 & 73.81 & 71.17 & \textbf{68.84} & 65.94 & 68.22 & 0.8133 & 0.9247   \\
        30  &   40    &   60  &  \underline{73.33} & \underline{74.54} & \textbf{71.32} & \underline{67.21} & \textbf{70.65} & \textbf{69.02} & \underline{0.8480} & \underline{0.9440}    \\
        30  &   40    &   80  &  \textbf{74.30} & \textbf{75.01} &  \underline{71.28} &   {67.17}  &     \textbf{70.65}    &   \textbf{69.02}  &   \textbf{0.8533}   &   \textbf{0.9451}   \\
        30  &   40    &   100  &  71.09 & 70.01 & 70.47 & 65.34 & \underline{66.50} & \underline{68.63} & 0.7435 & 0.8354    \\
        \bottomrule
        \end{tabular}
    \end{table*}
\paragrapht{Hidden dimension of autoencoders.}
    The encoder $g_\phi(\cdot)$ maps the frame token embeddings into the 256-dimensional latents, such that the encoder and decoder have 9.7M and 2.6M trainable parameters, respectively.
    To assess the impact of latent space size on performance, we train \method with various latent sizes and evaluate the performance on the Break Eggs dataset.
    \cref{tab:latent} summarizes the results, including the performance on downstream tasks and the number of trainable parameters corresponding to each latent size.
    Naturally, large latent spaces enhance representation capability but lead to more trainable parameters (e.g. 51.5M parameters with a 512-dimensional latent space for 12 encoder and 4 decoder layers) and require more extensive training data.
    The results indicate that increasing the latent size from 64 to 256 consistently improves performance.
    However, a further increase to a 512-dimensional latent space leads to performance degradation, attributed to the limited availability of training data.
    % Naturally, a larger latent size yields a higher representation capability while leading to more trainable parameters (e.g. 51.5M parameters with a 512-dimensional latent space for 12 encoder and 4 decoder layers) and requiring more training data.
    % As shown in \cref{tab:latent}, increasing the latent size from 64 to 256 achieves consistent performance improvement.
    % However, \method with a larger latent size of 512 shows poor performance due to insufficient training data.
    
\paragrapht{Token selection ratio.}
    Selective token merging (STM) allows \method to effectively capture action-related regions while excluding noisy regions without any training as shown in the main paper.
    We provide the performance of \method with various token selection ratios in the first panel of \cref{tab:ratio} and depict the selected tokens corresponding to each selection ratio in \cref{fig:stm}.
    The results show that the token selection ratio significantly affects the performance due to the difference in the field of view between ego and exo videos.
    In other words, a low selection ratio is insufficient to cover the action-related regions in ego videos (see \cref{fig:stma} and \cref{fig:stmb}), while a high selection ratio makes noisy tokens be included in exo videos (see \cref{fig:stme}).
    To balance the lack of information in the ego video and the unnecessary noise in the exo video, we set the token selection ratio to 0.3.

\paragrapht{Masking ratio.}
    We validate the effectiveness of the masking ratio in masked self-view modeling (MSM) and masked cross-view modeling (MCM) in the second and third panels of \cref{tab:ratio}.
    In MSM, we can guess a low masking ratio enables the model to easily solve each masked modeling problem, leading to insufficient causality learning.
    In practice, the results show significant performance drops in phase progression and Kendall's $\tau$.
    An extremely high masking ratio in MSM  makes learning the causality between frames hard as the decoder takes only a few clean tokens (or only masked tokens with a 100\% masking ratio).
    The low masking ratio in MCM degrades performance for a similar reason as in MSM.
    Meanwhile, a high masking ratio in MCM makes the masked cross-view modeling significantly difficult to solve with limited training data, showing performance drops across all downstream tasks.
    \method trained with the masking ratio of 0.4 and 0.8 in MSM and MCM achieves to produce the effective fine-grained view-invariant video representations.

    \begin{figure*}[t]
      \centering
        {\includegraphics[width=1.0\linewidth]{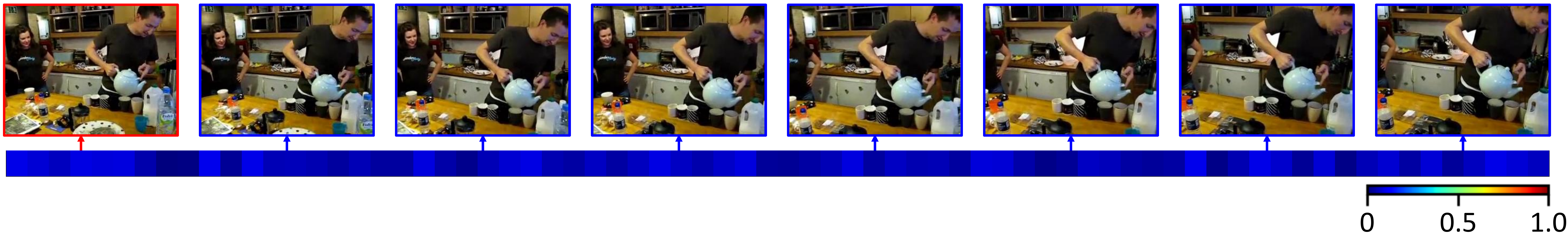}}\vspace{-7pt}       
      \caption{Visualization of the softmax similarity between final frame embeddings for a failure case from the \textit{Pour Liquid} benchmark. We depict the similarity score between only one reference token embedding (red box) and other token embeddings (blue boxes) for visibility.}\vspace{-3pt}
      \label{fig:atten}
    \end{figure*}

\subsection{Failure cases}
    While our \method significantly improves the performance across various benchmarks and experimental protocols, we observed that most failure cases occur in videos with slow movement transitions, particularly in exocentric videos.
    In such cases, frame embeddings tend to attend to each other uniformly, reducing the model's ability to capture meaningful temporal dependencies.
    \cref{fig:atten} illustrates a visualization of the softmax similarity score between the final frame embeddings for a failure case from the \textit{Pour Liquid} benchmark.
    Despite introducing positional embeddings and selective token merging, the embedding feature for a reference frame (red box) attends to all other embeddings similarly, resulting in less informative final representations.
    Beyond simple token selection of \method, learning-based token selection approach~\cite{run-length} may further improve the robustness of learned representations.

\section{Broader Impact}\label{sec:impact}

By achieving robust, view-invariant learning from unpaired ego-exo videos, \method can significantly advance the ability of AI to understand human actions and interactions across diverse perspectives, contributing to a wide range of real-world applications such as robotics, augmented and virtual reality, and assistive technologies.
Moreover, this research can facilitate new related research as follows;
\begin{itemize}
    \item Cross-view video generation: The video representations learned by \method contain fine-grained action context. In addition, the decoders used during training show a high recovery rate. This shows that it is possible to generate videos across views, which can be used to generate educational or instructional videos.
    \item Multi-view activity tracking: The view-consistent representations can be used in continuously tracking a person or object across various camera views (ego and exo) to maintain consistent identity and action recognition across perspectives, useful for applications in security and autonomous vehicles.
\end{itemize}
% \input{arXiv/Appendix_E}

% WARNING: do not forget to delete the supplementary pages from your submission 
% \input{sec/X_suppl}

\end{document}